\documentclass{article}
\usepackage{microtype}
\usepackage{graphicx}
\usepackage{subfigure}
\usepackage{booktabs} % for professional tables

\usepackage{hyperref}

\usepackage{graphicx}
\usepackage{subfigure}
\usepackage{booktabs} 
\usepackage{url}
\usepackage{amsfonts}
\usepackage{dsfont}
\usepackage{nicefrac}
\usepackage{amsmath}
\usepackage{amsthm}
\usepackage{amssymb}
\usepackage{stmaryrd}
\usepackage{amsmath,amssymb}
\usepackage{bbm}
\usepackage{dsfont}

\usepackage{graphicx}
\usepackage{subfigure} 
\newtheorem{theorem}{Theorem}

\newtheorem{definition}[theorem]{Definition}

\newtheorem{remark}[theorem]{Remark}

\numberwithin{theorem}{section}

\newcommand\disp{\mathrm{disp}}
\usepackage{amsmath,amssymb}
\DeclareMathOperator{\E}{\mathbb{E}}
\usepackage{amsthm}

\usepackage[subtle]{savetrees}

\newcommand{\eq}[1]{Eq.\:\eqref{#1}}

\usepackage{microtype}
\usepackage{graphicx}
\usepackage{subfigure}
\usepackage{booktabs} % for professional tables

\usepackage[flushleft]{threeparttable}

\usepackage{footnote}
\renewcommand*{\thefootnote}{\fnsymbol{footnote}}
\usepackage[dvipsnames]{xcolor}
\usepackage{float}

\usepackage{multirow}

\usepackage[font=small,labelfont=bf]{caption}

\usepackage[accepted]{icml2020_style/icml2020}

\icmltitlerunning{Implicit Class-Conditioned Domain Alignment for Unsupervised Domain Adaptation}

\begin{document}

\twocolumn[
\icmltitle{Implicit Class-Conditioned Domain Alignment\\ for Unsupervised Domain Adaptation}




\begin{icmlauthorlist}
\icmlauthor{Xiang Jiang}{imagia,dal}
\icmlauthor{Qicheng Lao}{imagia,mila}
\icmlauthor{Stan Matwin}{dal,polish}
\icmlauthor{Mohammad Havaei}{imagia}
\end{icmlauthorlist}

\icmlaffiliation{imagia}{Imagia, Canada}
\icmlaffiliation{dal}{Dalhousie University, Canada}
\icmlaffiliation{polish}{Polish Academy of Sciences, Poland}
\icmlaffiliation{mila}{Mila, Universit\'e de Montr\'eal, Canada}

\icmlcorrespondingauthor{Xiang Jiang}{xiang.jiang@dal.ca}

\icmlkeywords{domain adaptation, semi-supervised learning}

\vskip 0.3in
]



\printAffiliationsAndNotice{} % otherwise use the standard text.

\begin{abstract}
We present an approach for unsupervised domain adaptation---with a strong focus on practical considerations of within-domain class imbalance and between-domain class distribution shift---from a class-conditioned domain alignment perspective.
Current methods for class-conditioned domain alignment aim to \emph{explicitly} minimize a loss function based on pseudo-label estimations of the target domain.
However, these methods suffer from pseudo-label bias in the form of error accumulation.
We propose a method that removes the need for \emph{explicit} optimization of model parameters from pseudo-labels directly. Instead, we present a sampling-based \emph{implicit} alignment approach, where the sample selection procedure is \emph{implicitly} guided by the pseudo-labels.
Theoretical analysis reveals the existence of a domain-discriminator shortcut in misaligned classes, which is addressed by the proposed implicit alignment approach to facilitate domain-adversarial learning.
Empirical results and ablation studies confirm the effectiveness of the proposed approach, especially in the presence of within-domain class imbalance and between-domain class distribution shift.
\end{abstract}

\section{Introduction}
Supervised learning aims to extract statistical patterns from data by learning to approximate the conditional density $p(y|x)$. However, the generalization of the approximation is often sensitive to some dataset-specific factors.
Dataset shift~\cite{quionero2009dataset} frequently arises from real-world applications and can manifest in many different ways, such as sample selection bias~\cite{heckman1979sample,torralba2011unbiased}, class distribution shift~\cite{webb2005application}, and covariate shift~\cite{shimodaira2000improving}.
Unsupervised Domain Adaptation~(UDA) aims to address domain shift with access to labeled data in the source domain and unlabeled data in the target domain~\cite{pan2009survey}.
The fundamental algorithmic issue is to infer domain-invariant representations.% thereby achieving domain-agnostic classification.

While considerable progress has been made in UDA~\cite{ganin2016domain}, they tend to focus on marginal distribution matching in the feature space, and less emphasis is made on discovering label distributions.
In real-world applications, it is very common to have class imbalance within each domain and class distribution shift between different domains, necessitating the incorporation of label space distribution into adaptation.
\emph{Explicit} class-conditioned domain alignment~\cite{xie2018learning,pan2019transferrable,liang2019distant,deng2019cluster} has emerged as a key approach to promoting class-conditioned invariance by aligning prototypical representations of each class.
While explicit alignment has the advantage of directly minimizing class-conditioned misalignment, it presents critical vulnerabilities to error accumulation~\cite{chen2019progressive} and ill-calibrated probabilities~\cite{guo2017calibration} due to its dependence on \emph{explicit} supervision from pseudo-labels provided by model predictions.

We propose \emph{Implicit} Class-Conditioned Domain Alignment that removes the need for explicit pseudo-label based optimization.
Instead, we use the pseudo-labels \emph{implicitly} to \emph{sample} class-conditioned data in a way that maximally aligns the joint distribution between features and labels.
The primary advantage of the sampling-based implicit domain alignment is the ability to address within-domain class imbalance and between-domain class distribution shift, in addition to many other benefits such as applications in cost-sensitive learning.
The proposed method is simple, effective, and is supported by theoretical analysis on the empirical estimations of domain divergence measures.
It also overcomes limitations of explicit alignment by allowing the domain adaptation algorithm to discover class-conditioned domain-invariance in an unsupervised way without explicit supervision from pseudo-labels.

The contributions of this paper are as follows:~(i)~We propose implicit class-conditioned domain alignment to address the challenge of within-domain class imbalance and between-domain class distribution shift, which overcomes the limitation of error accumulation in explicit domain alignment; (ii)~We provide theoretical analysis on the empirical domain divergence and reveal the existence of a shortcut function that interferes with domain-invariant learning, which is addressed by the proposed approach; (iii)~We show that the proposed approach is orthogonal to the choice of domain adaptation algorithms and offers consistent improvements to two adversarial domain adaptation algorithms; (iv)~We report state-of-the-art UDA performance under extreme within-domain class imbalance and between-domain class distribution shift, and competitive results on standard UDA tasks.

\section{Preliminaries}
%In this section, we introduce unsupervised domain adaptation and review explicit class-conditioned domain alignment.
%\subsection{Problem Setup}
We follow the notations by~\cite{ben2010theory} and define a domain as an ordered pair consisting of a distribution $\mathcal{D}$ on the input space $\mathcal{X}$, and a labeling function $f:\mathcal{X}\to \mathcal{Y}$ that maps $\mathcal{X}$ to the label space $\mathcal{Y}$.
The source and target domains are denoted by $\langle \mathcal{D}_S, f_S\rangle$ and $\langle \mathcal{D}_T, f_T\rangle$, respectively.

%In unsupervised domain adaptation, the model is trained on labeled samples $\left \{(x_i,f_S(x_i))\right \}_{i=1}^{n}$ from the source domain where $x_i\sim \mathcal{D}_S$, together with unlabeled samples $\left \{x_j\right \}_{j=1}^{m}$ from the target distribution where $x_j\sim \mathcal{D}_T$.
In unsupervised domain adaptation, the model is trained on labeled data from the source domain, together with unlabeled data from the target domain.
The goal is to obtain a model $h\in\mathcal{H}$ which learns domain-invariant representations while simultaneously minimizing the classification error on $\mathcal{D}_S$.

%\subsection{Adversarial Domain Adaptation}
Adversarial training is the prevailing approach for domain adaptation~\cite{ganin2016domain}.
It formulates a minimax problem where the maximizer maximizes the estimation of the domain divergence between the empirical samples, and the minimizer minimizes the sum of the source error and the domain divergence estimation obtained from the maximizer.
%classifier $f$ and feature extractor $\phi$ are trying to minimize the source error $\epsilon_{\phi, f}(\hat{\mathcal{S}})$ and some domain divergence measure $d$, while the model $\theta$ is trying to maximize the estimation of the domain divergence between the empirical samples $\hat{\mathcal{S}}$ and $\hat{\mathcal{T}}$.
%The estimation of domain divergence can take many different forms, such as using a domain discriminator that predicts binary outcomes about whether the data is coming from the source or target distribution~\cite{ganin2016domain}, and estimating the discrepancies of two classifiers~\cite{saito2018maximum,pmlr-v97-zhang19i}.

%\subsection{Explicit Class-Conditioned Domain Alignment}
While matching the marginal distribution is a good step towards domain-invariant learning, it is still susceptible to the problem of conditional distribution mismatching.
Prototype-based class-conditioned domain alignment~\cite{luo2017label,xie2018learning,chen2019progressive,pan2019transferrable,liang2019distant,liang2019exploring} is designed to address this problem.
We refer to this group of methods as  \emph{explicit} class-conditioned domain alignment.
%, which aims to align the class-conditioned prototypical representations of each class between the source and target domains.
The explicit alignment is achieved by incorporating an auxiliary loss that minimizes the Euclidean distance of the class-conditioned prototypical representations ${\bf c}_j$ between the source and target domains.
The class-conditioned prototype ${\bf c}_j$ is the average representation for all examples in a domain with class label $j$.
% that is defined as ${\bf c}_j  = \frac{1}{N_j} \sum_{({\bf x}_i,y_i) \in \mathcal{D}} \mathbbm{1}_{\left \{ y_{i} = j \right \}} f_{\phi}({\bf x}_i)$, where $N_j$ is the number of examples for this class.

%For each class $t$ in the source and target domains, the model constructs prototypical representation ${\bf c}_t  = \frac{1}{K} \sum_{({\bf x}_i,y_i) \in \mathcal{D}} \mathbbm{1}_{\left \{ y_{i} = t \right \}} f_{\phi}({\bf x}_i)$ , and uses a loss function to measure the discrepancy between the class-conditioned prototypes.
%The minimax optimization problem is defined as the following:
%\begin{equation}
%\begin{split}
%    &\min_{f,\phi}\quad \epsilon_{\phi,f}(\hat{\mathcal{S}})+d_{\phi,\theta}(\hat{\mathcal{S}}, \hat{\mathcal{T}}) + \lambda \mathcal{L}_{\phi}(\hat{\mathcal{S}}, \hat{\mathcal{T}})\\
%    &\max_{\theta}\quad d_{\phi,\theta}(\hat{\mathcal{S}}, \hat{\mathcal{T}})
%\end{split}
%\end{equation}
%where $\lambda$ is a weighting factor.
%Due to the lack of labeled information in the target domain, explicit alignment relies on model predictions as pseudo-labels for~$\mathcal{L}_{\phi}(\hat{\mathcal{S}}, \hat{\mathcal{T}})$.

The main limitation of explicit class-conditioned domain alignment is in its reliance on explicit optimization of model parameters based on pseudo-labels.
% when estimating the label information of the target domain.
%We name this type of approach explicit alignment as the model is directly optimizing the loss based on the predicted pseudolabels.
This learning procedure is vulnerable to error accumulation~\cite{chen2019progressive} as mistakes in the pseudo-label predictions can gradually accumulate leading to poor local minima in EM-style training.
Furthermore, the pseudo-labels are likely to suffer from ill-calibrated probabilities~\cite{guo2017calibration}, especially for deep learning methods, which exacerbate the critical problem of error accumulation with misleadingly confident mistakes.
%and thus could give misleading confident mistakes which exacerbate the critical problem of error accumulation.
\section{Method}
We begin with theoretical motivations of implicit alignment by decomposing the empirical domain divergence measure into class-aligned and class-misaligned divergence, and show that misaligned divergence is detrimental to domain adaptation.
We then present the proposed implicit domain alignment framework that addresses class-misalignment.
%We proposed implicit class-conditioned domain alignment in order to address class-misaligned divergence.
% during the minimax optimization
%We then introduce the implicit domain alignment framework that aims to minimize the impact of the class-misaligned divergence.
%This motivate our proposed implicit alignment framework that minimizes empirical class misalignment.

\subsection{Theoretical Motivations}
\label{sec:theoretical_motivations}
%From theoretical analysis, we show our method gives a better empirical measure of domain divergence, and prevents the domain discriminator from learning class shortcuts.
%This motivation our proposed implicit alignment that aims to minimize the degree of class misalignment thereby achieving a better empirical measure of domain divergence.
%This prevents the domain discriminator from learning a shortcut that does not contribute to the domain adversarial training during the minimax optimization.
%We further show that, for the class-misaligned divergence, the domain discriminator could be confused with the disjoint label space between the empirical samples in the source and target domains, which creates a shortcut for the optimization of the domain discriminator and is detrimental to adversarial learning.
The $\mathcal{H}\Delta\mathcal{H}$ divergence between two domains is defined as
\begin{equation}
d_{\mathcal{H}\Delta\mathcal{H}}(\mathcal{D}_S, \mathcal{D}_T)=2\sup_{h,h'\in \mathcal{H}}\vert \E_{\mathcal{D}_T}  \left [ h\neq h' \right ]-  \E_{\mathcal{D}_S}\left [ h\neq h' \right ]\vert,
\end{equation}
where $\mathcal{H}$ denotes some hypothesis space, and $h\neq h'$ is the abbreviation for $h(x)\neq h'(x)$.
\citep{ben2010theory} theorized that the target domain error $\epsilon_T(h)$ is bounded by the error of the source domain $\epsilon_S(h)$ and the empirical domain divergence $\hat{d}_{\mathcal{H}\Delta\mathcal{H}}(\mathcal{U}_S, \mathcal{U}_T)$ where $\mathcal{U}_S$, $\mathcal{U}_T$ are unlabeled empirical samples drawn from $\mathcal{D}_S$, $\mathcal{D}_T$.%, and some constants about the hypothesis space $\mathcal{H}$ and sample size of $\mathcal{U}$.

\iffalse
\begin{theorem}[Bound of the target domain error~\citep{ben2010theory}]
\label{theorem:bound}
Let $\mathcal{H}$ be a hypothesis space of VC dimension $d$. If $\mathcal{U}_S$, $\mathcal{U}_T$ are unlabeled samples of size $m'$ each, drawn from $\mathcal{D}_S$, $\mathcal{D}_T$, respectively, then for any $\delta\in (0,1)$, with probability at least $1-\delta$~(over the choice of samples), for every $h\in\mathcal{H}$:
\begin{align}
    \epsilon_T(h)\leq &\epsilon_S(h)+\frac{1}{2}\hat{d}_{\mathcal{H}\Delta\mathcal{H}}(\mathcal{U}_S, \mathcal{U}_T)\\&+4\sqrt{\frac{2d\log(2m')+\log(\frac{2}{\delta})}{m'}} + \lambda
\end{align}
\end{theorem}
where $\epsilon_S(h)$ is the error on the source domain, $\hat{d}_{\mathcal{H}\Delta\mathcal{H}}(\mathcal{U}_S, \mathcal{U}_T)$ is the empirical estimation of the symmetric difference hypothesis space $d_{\mathcal{H}\Delta\mathcal{H}}(\mathcal{D}_S,\mathcal{D}_T)$, $\lambda$ is the ideal joint expected loss $\lambda=\arg \min_{h\in\mathcal{H}}\epsilon_S(h) +\epsilon_S(h)$, which is often assumed to be negligible.
%If $\lambda$ is large, it means the hypothesis space does not contain a good hypothesis $h\in\mathcal{H}$ with small errors on both domains~(which defeats the purpose of domain adaptation, or we need a better architecture).
\fi
%In deep neural networks, learning is accomplished by minibatch-based stochastic optimization where only limited data are available at each optimization step.
In deep learning, minibatch-based optimization limits the amount of data available at each training step.
This necessitates the analysis of the empirical estimations of  $d_{\mathcal{H}\Delta\mathcal{H}}$ at the minibatch level, so as to shed light on the learning dynamics.
%Implicit domain alignment plays an important role in the empirical estimations of  $d_{\mathcal{H}\Delta\mathcal{H}}(\mathcal{D}_S,\mathcal{D}_T)$, where only finite samples $\mathcal{U}_S$, $\mathcal{U}_T$ are available.
%Below we define the empirical $\mathcal{H}\Delta\mathcal{H}$ divergence on mini-batches.% which will be used for the construction of our main theorem.
\begin{definition}
Let $\mathcal{B}_S$, $\mathcal{B}_T$ be minibatches from $\mathcal{U}_S$ and $\mathcal{U}_T$, respectively, where $\mathcal{B}_S\subseteq \mathcal{U}_S$, $\mathcal{B}_T\subseteq \mathcal{U}_T$, and $\vert \mathcal{B}_S\vert=\vert\mathcal{B}_T\vert$.
The empirical estimation of $d_{\mathcal{H}\Delta\mathcal{H}}(\mathcal{B}_S, \mathcal{B}_T)$ over the minibatches $\mathcal{B}_S$, $\mathcal{B}_T$ is defined as
\begin{align}
\hat{d}_{\mathcal{H}\Delta\mathcal{H}}(\mathcal{B}_S, \mathcal{B}_T)
=\sup_{h,h'\in \mathcal{H}}\left \vert \sum_{\mathcal{B}_T}\left [ h\neq h' \right ]-  \sum_{\mathcal{B}_S}\left [ h\neq h' \right ]\right \vert.
\end{align}
\end{definition}
%For simplicity, we drop the multiple $\frac{1}{m_b}$ in the following analysis as it does not affect the result of optimization.

\begin{theorem}[The decomposition of $\hat{d}_{\mathcal{H}\Delta\mathcal{H}}(\mathcal{B}_S, \mathcal{B}_T)$]
\label{theorem:bound}
Let $\mathcal{H}$ be a hypothesis space and $\mathcal{Y}$ be the label space of the classification task where $\mathcal{B}_S$, $\mathcal{B}_T$ are minibatches drawn from $\mathcal{U}_S$, $\mathcal{U}_T$, respectively, and $Y_S$, $Y_T$ are the label set of $\mathcal{B}_S$, $\mathcal{B}_T$. We define three disjoint sets on the label space: the shared labels $Y_C:=Y_S \cap  Y_T$, and the domain-specific labels $\overline{Y_{S}}:=Y_S-Y_C$, and $\overline{Y_{T}}:=Y_T-Y_C$.
We also define the following disjoint sets on the input space where $\mathcal{B}_S^C:=\left \{ x\in \mathcal{B}_S \mid y\in Y_C \right \}$, $\mathcal{B}^{\overline{C}}_{S}:=\left \{ x\in \mathcal{B}_S \mid y\notin Y_C \right \}$, $\mathcal{B}_T^C:=\left \{ x\in \mathcal{B}_T \mid y\in Y_C \right \}$, $\mathcal{B}^{\overline{C}}_{T}:=\left \{ x\in \mathcal{B}_T \mid y\notin Y_C \right \}$. 
The empirical $\hat{d}_{\mathcal{H}\Delta\mathcal{H}}(\mathcal{B}_S, \mathcal{B}_T)$ divergence can be decomposed into class aligned divergence and class-misaligned divergence:
\begin{equation}
    \hat{d}_{\mathcal{H}\Delta\mathcal{H}}(\mathcal{B}_S, \mathcal{B}_T)=\sup_{h,h'\in \mathcal{H}}\left |\xi^C(h, h') +  \xi^{\overline{C}}(h, h') \right |,
\end{equation}
where 
\begin{align}
    \xi^C(h, h')= \sum_{\mathcal{B}_T^C}\mathbbm 1\left [ h\neq h' \right ]-  \sum_{ \mathcal{B}_S^C}\mathbbm 1\left [ h\neq h' \right ],\\
    \xi^{\overline{C}}(h, h')= \sum_{\mathcal{B}^{\overline{C}}_{T}}\mathbbm 1\left [ h\neq h' \right ]-  \sum_{\mathcal{B}^{\overline{C}}_{S}}\mathbbm 1\left [ h\neq h' \right ].
\end{align}
\end{theorem}
The proof is provided in supplementary materials.% by rewriting the summation over $\mathcal{B}$ into the sum of disjoint subsets $\mathcal{B}^C$ and $\mathcal{B}^D$.
%\newpage
\begin{figure}[!t]
    \centering
    \includegraphics[width=0.4\textwidth]{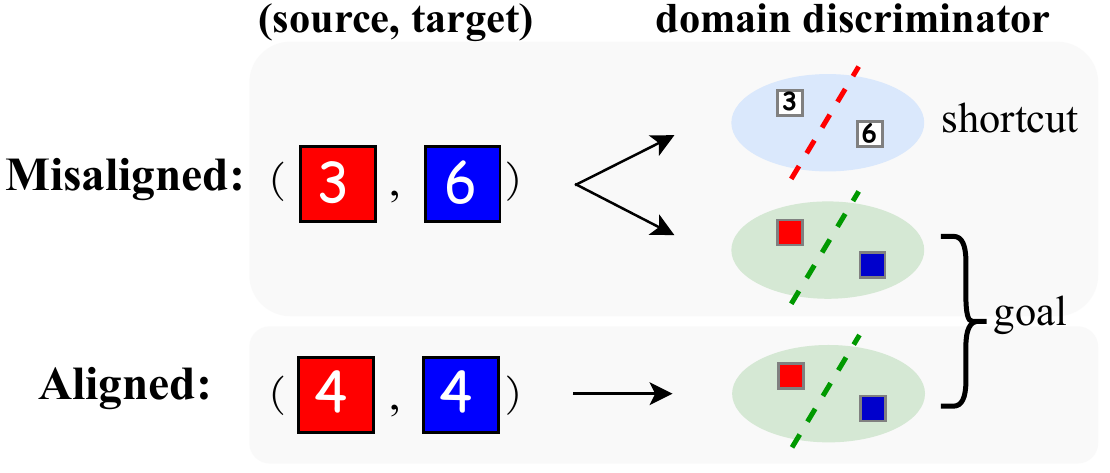}
    \vspace{-10pt}
    \caption{Illustration of the domain discriminator shortcut. The domain discriminator aims to distinguish between different domains~({\color{red}{red}} and {\color{blue}{blue}}), where the decision boundary is represented by dashed lines.
    But misaligned samples create a shortcut where the domain labels can be directly determined by the misaligned class labels~(3 and 6).
    The decision boundary of the resulting shortcut is independent of the covariate that causes the domain difference, which does not contribute to adversarial domain-invariant learning.
    }
    \label{fig:shortcut}
    \vspace{-12pt}
\end{figure}
\begin{remark}[The domain discriminator shortcut]
Let the ordered triple $\left ( x, y_c, y_d\right )$ denote data sample $x$, and its associating class label $y_c$ and domain label $y_d$, respectively, where $x\in\mathcal{B}$, $y_c\in Y$ and $y_d\in \{0,1\}$.
Let $f_c$ be a classifier that maps $x$ to a class label $y_c$.
Let $f_d$ be a domain discriminator that maps x to a binary domain label $y_d$.
For the empirical class-misaligned divergence $\xi^{\overline{C}}(h, h')$ with sample $x\in \mathcal{B}^{\overline{C}}_{S} \cup  \mathcal{B}^{\overline{C}}_{T}$, there exists a domain discriminator shortcut function
\begin{equation}
    f_d(x)=\left\{\begin{matrix}
1 & f_c(x)\in \overline{Y_{S}}\\ 
0 & f_c(x)\in \overline{Y_{T}}, 
\end{matrix}\right.
\end{equation}
such that the domain label can be solely determined by the domain-specific class labels.
This shortcut interferes with adversarial domain adaptation because the model could bypass the optimization for domain-invariant representations, but rather optimize for a shortcut function that is independent of the covariate contributing to the domain difference.
\end{remark}
Figure~\ref{fig:shortcut} illustrates a toy example where the source and target domains are aligned for class 4 but misaligned between classes {\color{red}3} and {\color{blue}6} as a result of random sampling in the minibatch construction.
The domain discriminator aims to predict domain labels based on their domain information, i.e., {\color{red}red} and {\color{blue}blue}.
However, due to the class shortcut for the misaligned samples~({\color{red}3} and {\color{blue}6}), the domain discriminator could infer domain labels based on class information directly~(digits 3 and 6), without the need to learn domain-specific information.
This problem of class-misalignment is especially pronounced under extreme within-domain class imbalance and between-domain class distribution shift, where a simple random sample is more likely to fail in providing good coverage of the label space.
\subsection{Implicit Class-Conditioned Domain Alignment}
\begin{figure}[!t]
    \centering
    \includegraphics[width=0.45\textwidth]{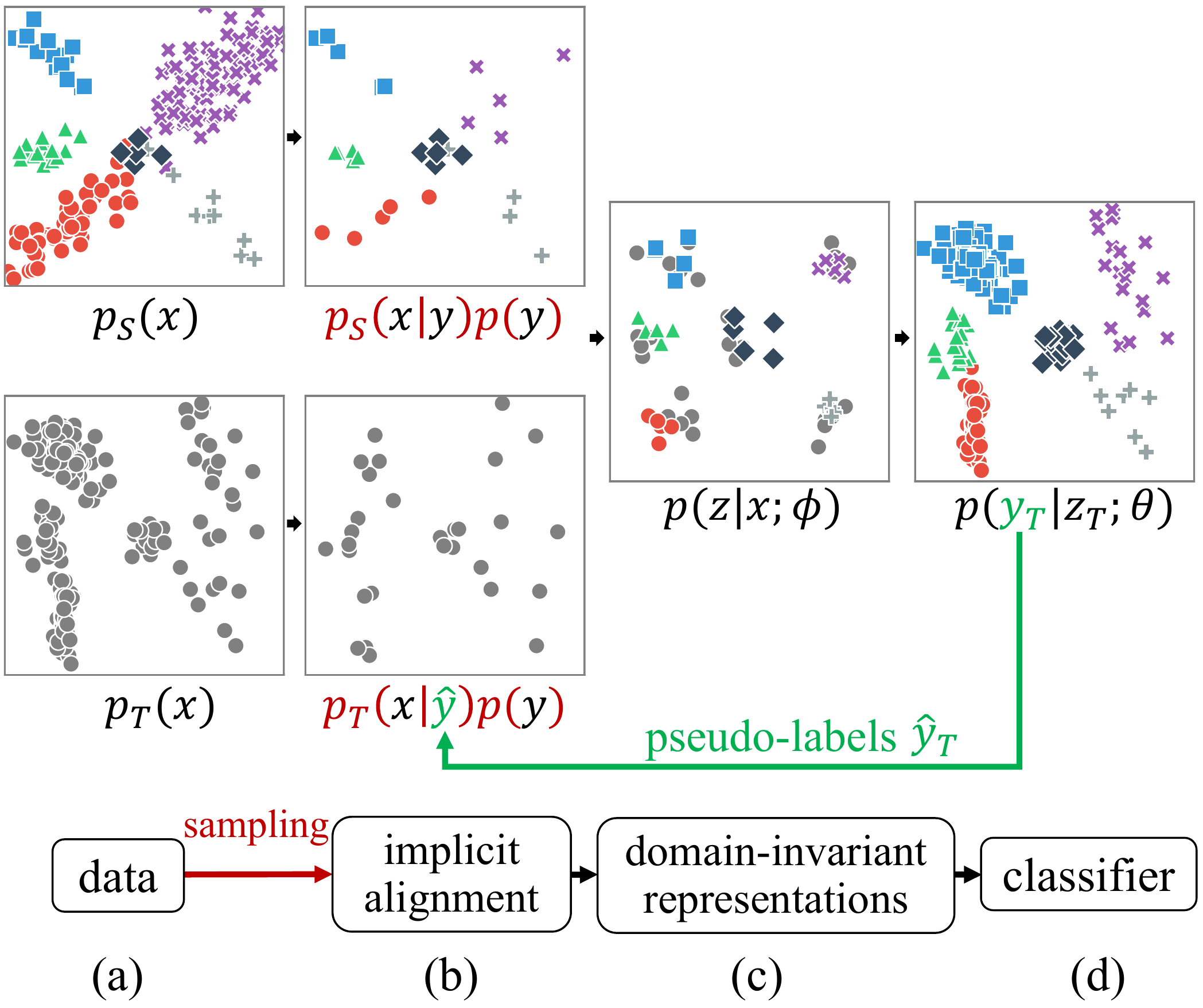}
    \vspace{-10pt}
    \caption{The proposed framework.
    (a)~We aim to align the source domain $p_S(x)$, colored by classes, with unlabeled target domain $p_T(x)$.
    %The model aligns the source and target domains in the label space by using the same 
    (b)~For $p_S(x)$, we sample $x\sim p_S(x\vert y)p(y)$ based on the \emph{alignment distribution} $p(y)$.
    For $p_T(x)$, we sample a \emph{class aligned} minibatch $x\sim p_T(x\vert \hat{y})p(y)$ using identical~$p(y)$, with the help of pseudo-labels $\hat{y}_T$. (c)~The adversarial training aims to acquire domain-invariant representations $z$ from the feature extractor parameterized by $\phi$.
    (d)~The classifier predicts class labels from $z$.}
    \label{fig:diagram}
    \vspace{-15pt}
\end{figure}
Having identified the domain discriminator shortcut in class misaligned empirical samples, we now propose framework that aligns the two domains from a sampling perspective.% providing a unified adversarial training procedure without the use of additional losses.% or hyper-parameters.

Figure~\ref{fig:diagram} depicts the proposed implicit class-conditioned domain alignment framework.
We aim to align $p_S(x)$ and $p_T(x)$ at the input and label space jointly with the factorization $p(x,y)=p(x|y)p(y)$ while ensuring that the sampled classes are aligned between the two domains.
The alignment distribution $p(y)$ is pre-specified, e.g., uniform distribution, to ensure samples are aligned in the shared label space in spite of different empirical label distributions of the two domains.
This implicit alignment procedure minimizes the class-misaligned divergence $\xi^{\overline{C}}(h, h')$, providing a more reliable empirical estimation of domain divergence.
For the unlabeled target domain, we use the model predictions to sample class-conditioned data from $p_T(x|\hat{y})$ to approximate $p_T(x|y)$.

\subsubsection{Class-Aligned Sampling Strategy}
\label{sec:minibatch-sampling-strategy}
%\emph{Overall sampling strategy.} 
Algorithm~\ref{alg:implicit} presents the proposed sampling procedure that selects class-aligned examples for minibatch training.
%It is a type of stratified random sampling where subgroups of the dataset are determined by their class labels.
It is a type of stratified sampling where the dataset is partitioned into mutually exclusive subgroups to reflect the label information in a class-aligned manner.
%Below we describe the algorithm.

First, we predict pseudo-labels of the target domain using the classifier $f_c(\cdot;\theta)$ parameterized by $\theta$.
The pseudo-labels will be later used in class-conditioned sampling.
Second, we sample a set $Y$ from the label space $\mathcal{Y}$ where $p(y)$ defines the probability with which we pick the classes to align so as to ensure the empirical samples of the source and target domains share the same $Y$.
This in turn minimizes the class-misaligned divergence $\xi^{\overline{C}}(h, h')$.
Third, for each class $y_i\in Y$, we sample class-conditioned examples for the source and target domains, respectively, and store them in $(X_S',Y_S')$ and $X_T'$.
This is equivalent to performing a table lookup to select a subset~$\mathcal{B}_i$ where all examples belong to class $y_i$, followed by random sampling in~$\mathcal{B}_i$.
We use pseudo-labels to sample the target domain due to the lack of ground-truth labels.
Once we obtained the class-aligned minibatch, we use it to train unsupervised domain adaptation algorithm and repeat this process until the model converges.

%This \emph{sampling} procedure aims to draw \emph{class-aligned} minibatches for stochastic optimization.
%The outermost plate in Figure~\ref{fig:plate} represents all the variables of a minibatch.
%$\alpha$ is the parameter that specifies which class to sample in the label space to be aligned, and $N$ is the number of classes.
%The inner plates represent $K$ class-conditioned samples of the source and target domains, respectively, and they are conditioned on the same class label $Y$ for alignment.
%Due to the lack of ground-truth labels, the target domain samples are also dependent on the model parameter $\theta$ for predicting the pseudo-labels.
%This sampling strategy ensures class alignment because samples in the source and target domains share the same label distribution defined by $\alpha$.
This algorithm addresses class imbalance within each domain as well as class distribution shift between different domains by specifying the sampling strategy $p(y)$ in the label space.
We use uniform sampling $p(y)$ for all experiments in this paper, and leave more advanced specifications and their applications to cost-sensitive domain adaptation as future work.
%\SetAlFnt{\small}
\begin{algorithm}[tb]
   \caption{The proposed implicit alignment training}
   \label{alg:example}
\begin{algorithmic}[1]
   \STATE {\bfseries Input:} dataset $S=\{(x_i, y_i)\}_{i=1}^{N}$, $T=\{x_i\}_{i=1}^{M}$,
   \STATE {~~~~~~~~~~~~~~~}label space $\mathcal{Y}$, label alignment distribution $p(y)$,
   \STATE {~~~~~~~~~~~~~~~}classifier $f_c(\cdot;\theta)$
   \WHILE{\NOT converged}
   \STATE \textsl{{\color{blue}{\# predict pseudo-labels for $T$}}}
   \STATE $\hat{T}\leftarrow\{(x_i,f_c(x_i;\theta))\}_{i=1}^{M}$ where $x_i\in T$
   %\newline
   \STATE \textsl{{\color{blue}{\# sample $N$ unique classes in the label space}}}
   \STATE $Y\leftarrow$ draw $N$ samples in $\mathcal{Y}$ from $p(y)$
   %\newline
   \STATE \textsl{{\color{blue}{\# sample $K$ examples conditioned on each $y_i\in Y$}}}
   \FOR{$y_i$ in $Y$}
   \STATE $(X_S',Y_S')\leftarrow$draw $K$ samples in $S$ from $p_S(x|y_i)$
   \STATE $X_T'\leftarrow$draw $K$ samples in $\hat{T}$ from $p_T(x|y_i)$
   \ENDFOR
   %\newline
   \STATE \textsl{{\color{blue}{\# domain adaptation training on this minibatch}}}
   \STATE train minibatch ($X_S'$,$Y_S'$,$X_T'$)
   \ENDWHILE
\end{algorithmic}
\label{alg:implicit}
\end{algorithm}

%\emph{Dealing with Data imbalance and Class Distribution Shift.}
%Data imbalance refers to different proportions of classes within one domain.
%Class distribution shift refers to the difference of class distribution between the source and target domains.
%Data imbalance and class distribution shift frequently arises from real-world applications.
%The proposed implicit class-conditioned alignment is able to address data imbalance and class distribution shift by specifying a proper prior distribution $\alpha$ when aligning the input space.
%If we employ uniform sampling in the label space, we obtain class-balanced samples in both domains, which naturally addresses the problem of class imbalance and class distribution shift.
%Additionally, for cost-sensitive learning, we can pre-specify $\alpha$ to suit our needs.
%We use uniform sampling for all experiments in this paper, and we leave the specification of the prior distribution as future work.

%\paragraph{Pseudo-label bias.}
%Pseudo-labels are biased towards distributions of the source domain, which tends to assign images that are similar to the source domain with high confidence.

%\paragraph{Prototypical networks do not capture multiple modes in the source domain.}

\subsubsection{Integrating Implicit Alignment into\\Classifier-Based Domain Discrepancy Measure}
Section~\ref{sec:minibatch-sampling-strategy} describes the implicit alignment algorithm from a sampling perspective, where we sample minibatches in a way that maximizes class alignment implicitly.
This sampling strategy is independent of the choice of domain divergence measures.
In this section, we show how to integrate the sampling approach into Margin Disparity Discrepancy~(MDD)~\cite{pmlr-v97-zhang19i}---a state-of-the-art classifier-based domain discrepancy measure---to further facilitate implicit alignment.
%improve the functional approximation of classifier-based domain discrepancy, as an extension to our implicit alignment theory.
%extend our theory into classifier-based domain divergence measures and demonstrate how to take our theory one step further to also improve the empirical classifier-based domain discrepancy estimations.
%\paragraph{Classifier-based domain discrepancy.}
MDD is defined as
	\begin{equation}
			 d_{f,\mathcal F}( S, T)=\sup_{f'\in\mathcal F}\Big{(}\disp_{\mathcal{D}_T}(f',f)-\disp_{\mathcal{D}_S}(f',f)\Big{)},
			 \label{eq:mdd-discrepancy}
	\end{equation}
where $f$ and $f'$ are two independent scoring functions that predict class probabilities, and $\disp(f',f)$ is a disparity measure between the scores provided by the classifiers $f'$ and $f$.
%where $f$ is a scoring function, i.e., classifier, that predicts probabilities of different classes, $f'\in\mathcal{F}$ is an auxiliary scoring function for estimating the domain divergence, and $\disp_{ T}^{(\rho)}$ denote some disparity measure between $f'$ and $f$.
The domain divergence is to estimate the discrepancy between the disparity measures of the two domains.
%This can be trained in an adversarial way similar to adversarial domain discriminators.
%Below we show how MDD suffers from class-misalignment and proposes a masking scheme on $f$ and $f'$ to address this problem.

%\emph{Class misalignment in classifier-based discrepancy.}
Following notations in Theorem~\ref{theorem:bound}, we define the empirical MDD on class-misaligned samples as
	\begin{equation}
			 \hat{d}_{f,\mathcal F}( \mathcal{B}^{\overline{C}}_{S}, \mathcal{B}^{\overline{C}}_{T})=\sup_{f'\in\mathcal F}\Big{(}\sum_{\mathcal{B}^{\overline{C}}_{T}}\disp(f',f)-\sum_{\mathcal{B}^{\overline{C}}_{S}}\disp(f',f)\Big{)}.
			 \label{eq:empirical-misaligned-classifier-divergence}
	\end{equation}
Because $\mathcal{B}^{\overline{C}}_{S}$ and $\mathcal{B}^{\overline{C}}_{T}$ are disjoint in the label space, there exists a shortcut solution% to the maximization problem in the following form
\begin{equation}
    \disp(f'(x),f(x))=\left\{\begin{matrix}
0 & f_c(x)\in \overline{Y_{S}}\\ 
1 & f_c(x)\in \overline{Y_{T}},
\end{matrix}\right.
\end{equation}
which maximizes the divergence estimation of~\eq{eq:empirical-misaligned-classifier-divergence}.
%This establishes a shortcut from the label space to the estimation of the classifier-based domain discrepancy without observing data in the input space.
%Furthermore, if we denote parameters of the auxiliary classifier that contributed to the prediction of $\overline{Y_{S}}$, $\overline{Y_{T}}$ as $f_P$ and $f_Q$, respectively.
%We can find different portions of the classifier's parameters are trained on 
%, but it is not a useful measure of domain divergence because it bypasses the covariate that contributes to the domain difference.
Although class-aligned sampling can mitigate this problem, it is difficult to fully eliminate the impact of misalignement due to imperfect pseudo-labels.
To further eliminate the detrimental impact of class-misalignment, we introduce a masking scheme on the scoring functions $f$ and $f'$ defined as
	\begin{equation}
	\begin{aligned}
	&\hat{d}_{f,\mathcal F}( \mathcal{B}_S, \mathcal{B}_T)
	\\&=\sup_{f'\in\mathcal F}\Big{(}\sum_{\mathcal{B}_T}\disp(f'\odot\omega,f\odot\omega)-\sum_{\mathcal{B}_S}\disp(f'\odot\omega,f\odot\omega)\Big{)},
	%\nonumber
	\end{aligned}
	\label{eq:classifier-mask}
	\end{equation}
where $f\odot \omega$ denotes element-wise multiplication between the output of $f$ and $\omega$.
The alignment mask $\omega$ is a binary vector that denotes whether the $i$-th class is present in the sampled classes $Y$~(i.e., the classes that we intend to align in the current minibatch).
By doing so, we simultaneously align the source and target domains (i) in the input space and (ii) in the functional approximations of the domain divergence by masking the scoring functions $f$ and $f'$.
%Note that the masking is applied in addition to the sampling approach described in Section~\ref{sec:minibatch-sampling-strategy}.
%This simple treatment allows us to further eliminate the detrimental impact of class-misalignment and provide more reliable empirical estimations of the classifier-based divergence estimation.
%We apply the same sampling procedure to align the input space, and the same $\omega$ is applied to both domains in \eq{eq:classifier-mask}.

\section{Experiments}
%Our PyTorch code is provided in \href{https://drive.google.com/open?id=1qw8tQNjOv1kokqkjVeww1yZkA7x3Tcde}{this anonymous link}.
%\vspace{-8pt}
\renewcommand{\thefootnote}{\fnsymbol{footnote}}
%\footnote[1]{\ast}

\begin{table*}[ht]
	%\addtolength{\tabcolsep}{-5pt}
	\centering
	\caption{Per-class average accuracy on Office-Home dataset with RS-UT label shifts (ResNet-50).}
	\label{table:officehome-imbalanced}
%	\vskip 0.05in
	\scalebox{0.75}{
	\begin{threeparttable}
		\begin{tabular}{lccccccc}
		\toprule
    Methods  &\textbf{ Rw}$\shortrightarrow$\textbf{Pr} & \textbf{Rw}$\shortrightarrow$\textbf{Cl} 
          & \textbf{Pr}$\shortrightarrow$\textbf{Rw} & \textbf{Pr}$\shortrightarrow$\textbf{Cl} & \textbf{Cl}$\shortrightarrow$\textbf{Rw} &\textbf{Cl}$\shortrightarrow$\textbf{Pr} &  Avg  \\
    \midrule
    Source Only$^{\color{blue}\dagger}$ & 69.77 & 38.35 & 67.31 & 35.84 & 53.31 & 52.27 & 52.81  \\
    \midrule
    BSP~\cite{chen2019transferability}$^{\color{blue}{\dagger}}$ & 72.80  & 23.82 & 66.19 & 20.05 & 32.59 & 30.36 & 40.97  \\
    PADA~\cite{cao2018partial}$^{\color{blue}{\dagger}}$ & 60.77  & 32.28  & 57.09  & 26.76  & 40.71  & 38.34  & 42.66  \\
    BBSE~\cite{lipton2018detecting}$^{\color{blue}{\dagger}}$ & 61.10  & 33.27  & 62.66  & 31.15  & 39.70  & 38.08  & 44.33  \\
    MCD~\cite{saito2018maximum}$^{\color{blue}{\dagger}}$ & 66.03  & 33.17  & 62.95  & 29.99  & 44.47  & 39.01  & 45.94  \\
    DAN~\cite{long2015learning}$^{\color{blue}{\dagger}}$ & 69.35  & 40.84  & 66.93  & 34.66  & 53.55  & 52.09  & 52.90  \\
    F-DANN~\cite{wu2019domain}$^{\color{blue}{\dagger}}$ & 68.56  & 40.57  & 67.32  & 37.33  & 55.84  & 53.67  & 53.88  \\
    JAN~\cite{long2017deep}$^{\color{blue}{\dagger}}$ & 67.20  & 43.60  & 68.87  & 39.21  & 57.98 & 48.57  & 54.24  \\
    DANN~\cite{ganin2016domain}$^{\color{blue}{\dagger}}$ & 71.62  & 46.51  & 68.40  & 38.07  & 58.83  & 58.05 & 56.91  \\
    MDD (random sampler) & 71.21 & 44.78 & 69.31 & 42.56 & 52.10 & 52.70 & 55.44 \\
    MDD (source-balanced sampler) & 76.06 & 47.38 & 71.56 & 40.03 & 57.46 & 58.54 & 58.50 \\
    COAL~\citep{tan2019generalized}$^{{\color{blue}{\dagger}}\color{black}{,}{\color{red}{\ddagger}}}$ & 73.65 & 42.58 & 73.26 & 40.61 & 59.22 & 57.33 & 58.40 \\%\midrule
    
    MDD+Explicit Alignment~(basic)$^{\color{red}{\ddagger}}$& 69.52  & 44.70 & 69.59 & 40.27 & 53.02 & 53.39 & 55.08 \\
    MDD+Explicit Alignment~(moving avg.)$^{\color{red}{\ddagger}}$ & 71.37 & 45.26 & 69.69 & 40.28 & 52.92 & 52.69 & 55.37 \\
    MDD+Explicit Alignment~(curriculum)$^{\color{red}{\ddagger}}$ & 70.02 & 45.48 & 69.71 & 40.86 & 53.26 & 52.99 & 55.39 \\
    \textbf{MDD+Implicit Alignment} & \textbf{76.08}  & \textbf{50.04} & \textbf{74.21} & \textbf{45.38} & \textbf{61.15} & \textbf{63.15} & \textbf{61.67} \\
    \bottomrule%[1.2pt]
    
    \end{tabular}%
    \begin{tablenotes}
    \item[${\color{blue}{\dagger}}$] \emph{Source}: Data of these baseline methods are cited from~\cite{tan2019generalized}.
    \item[$\color{red}{\ddagger}$] Methods using explicit class-conditioned domain alignment.
  \end{tablenotes}
    \end{threeparttable}}
	\vspace{-10pt}
\end{table*}
%We validate the impact of implicit alignment and its robustness to class distribution shift and pseudo-label bias.
%We evaluate our model on two popular domain adaptation datasets, and show the impact of implicit alignment and its robustness to data imbalance and class distribution shift.
%The code is provided in supplementary materials.

\subsection{Setup}
\footnotetext{\scriptsize {\color{red}Code}: \url{https://github.com/xiangdal/implicit_alignment} }
%\subsubsection{Datasets}
%We use the standard Office-31~\cite{saenko2010adapting} and three different versions of Office-Home~\cite{venkateswara2017deep} in our experiments.
%Note that the domain gap in Office-Home is greater than the domain gap in Office-31.
\textbf{Datasets.}
We evaluate on Office-31, Office-Home and VisDA2017.
Office-31~\cite{saenko2010adapting} has three domains~(\textbf{A}mazon, \textbf{D}SLR and \textbf{W}ebcam)~with 31 classes.
%Office-Home~\cite{venkateswara2017deep} contains four domains~(\textbf{Ar}t, \textbf{Cl}ip Art, \textbf{Pr}duct, and \textbf{Re}al-world)~with 65  classes.
We use three versions of Office-Home~\cite{venkateswara2017deep} that contains four domains~(\textbf{Ar}t, \textbf{Cl}ip Art, \textbf{Pr}duct, and \textbf{R}eal-\textbf{w}orld)~with 65 classes:
(i)~``standard'': the standard Office-Home dataset.
(ii)~``balanced''~\cite{tan2019generalized}: a subset of the standard dataset where each class has the same number of examples.
(iii)~``RS-UT'': Reversely-unbalanced Source (RS) and Unbalanced-Target (UT) distribution~\cite{tan2019generalized} where both domains are imbalanced, but the majority class in the source domain is the minority class in the target domain.
VisDA2017~(synthetic$\rightarrow$real)~\cite{peng2017visda} is a large-scale dataset with 12 classes and more than 200k images.
%Further dataset details are in the supplementary materials.

\textbf{Model architecture.}
We use ResNet-50~\cite{he2016deep} pre-trained from ImageNet~\cite{russakovsky2015imagenet} as the backbone, and use hyper-parameters from~\cite{pmlr-v97-zhang19i} for MDD-based domain discrepancy measure.
The batch size is 31 for Office-31 and 50 for Office-Home.

\textbf{Baselines.}
Our main explicit alignment baselines are COAL~\cite{tan2019generalized}, PACET~\cite{liang2019exploring} and MCS~\cite{liang2019distant}, state-of-the-art explicit alignment methods based on domain discriminator discrepancy.
%For the imbalanced experiments, our main baseline is COAL~\cite{tan2019generalized}, an explicit alignment method designed for class imbalance and distribution shift.
%For the experiments with the standard datasets, our main baselines are PACET~\cite{liang2019exploring} and MCS~\cite{liang2019distant}, state-of-the-art explicit alignment algorithms based on domain discriminator-based discrepancy measures.
As our domain discrepancy measure is MDD, we re-implement various MDD-based explicit alignment for fair comparison.

\textbf{Computational efficiency.}
%Self-training requires the estimation of target domain labels, which could be time-consuming depending on the size of the target domain.
% As an example, it takes 14 seconds to get the predicted labels on the Clipart domain for office-home, whereas it takes less than a second to train one batch.
We only update pseudo-labels periodically, i.e., every 20 steps, instead of at every training step.
%To minimize the efficiency impact of self-training, we update the pseudo-labels every 20 training steps.
We show in the supplementary materials that our method does not require more frequent pseudo-label updates.% as they exhibit similar performance on the target domain. %(show in supplementary materials)
%We leave the caching and updating strategies of pseudo-labels as future work.

\subsection{Evaluating Extreme Class Distribution Shift}
\begin{figure}
    \centering
    \includegraphics[width=0.4\textwidth]{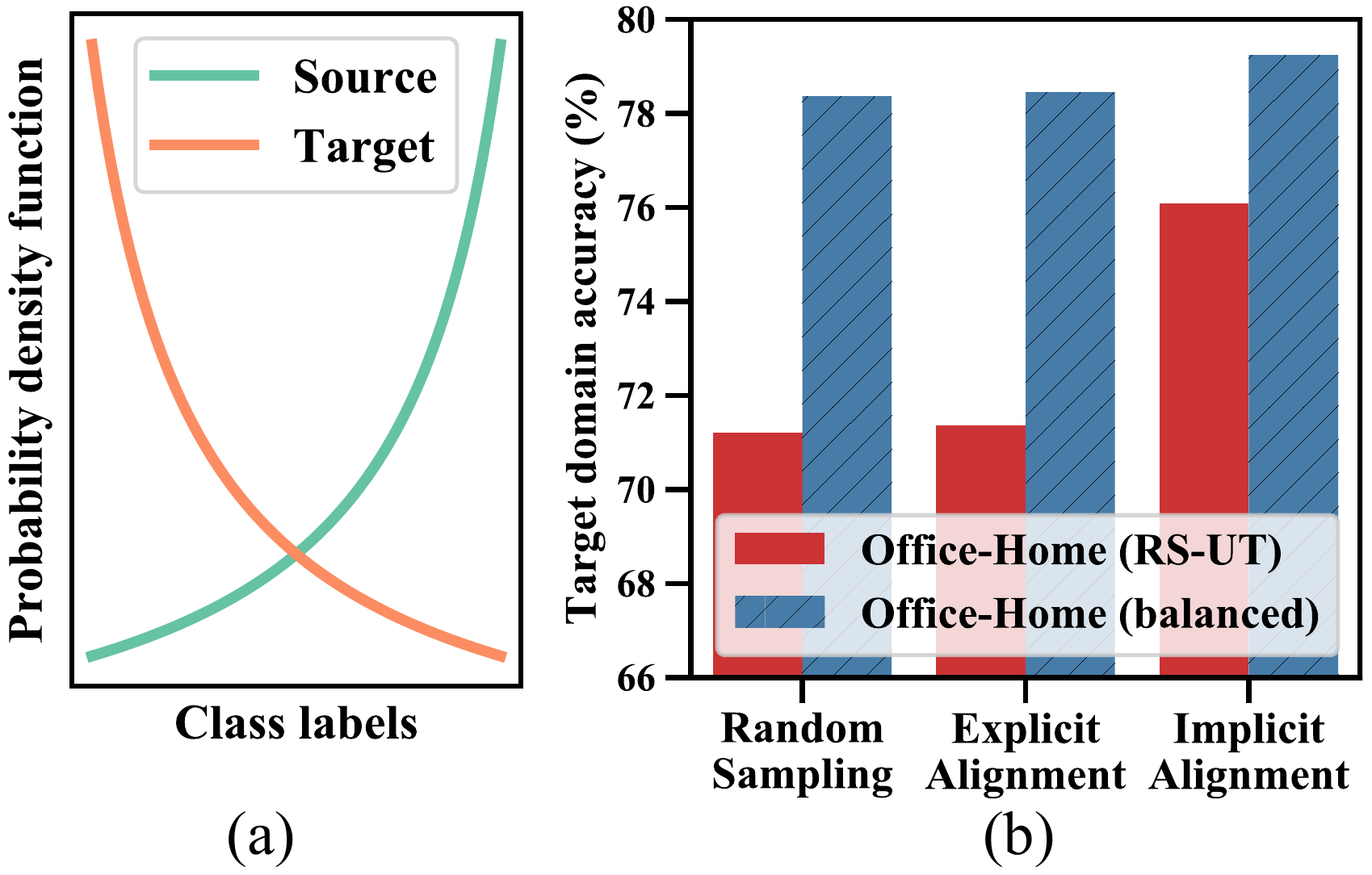}
    \vspace{-12pt}
    \caption{(a)~Source and target class distribution of Office-Home (RS-UT). (b)~Accuracy comparison between Office-Home~(RS-UT) and Office-Home~(balanced) for Rw$\rightarrow$Pr. %This shows implicit alignment is more robust to label imbalance, hence  a smaller gap.
    }
    \label{fig:imbalance_accuracy_gap}
    \vspace{-18pt}
\end{figure}

%\subsection{Results and Discussions}
%\subsubsection{Extreme Class Distribution Shift}
\label{sec:extreme-class-distribution-shift}
%Current work on domain adaptation mainly focuses on the classification accuracy as evaluation measure, but does not take account into the imbalanceness in the label space. This is not desirable as a model with high classification accuracy might merely works better on the majority classes while fails at minority classes. 
We use Office-Home~(RS-UT), described in Figure~\ref{fig:imbalance_accuracy_gap}~(a), to evaluate the performance of different methods under extreme within-domain class imbalance and between-domain class distribution shift where the majority classes in the source domain are minority classes in the target domain.
Table~\ref{table:officehome-imbalanced} presents the per-class average accuracy on Office-Home~(RS-UT).
Our main baseline is the explicit alignment method ``covariate and label shift co-alignment''~(COAL) designed to address data imbalance and class distribution shift.
Our proposed implicit domain alignment works the best.

\subsubsection{The impact of class distribution shift}
Many baseline methods suffer from class distribution shift, and their performances degrade to ``Source Only'' training as they do not take into account within-domain class imbalance and between-domain class distribution shift.
For MDD-based methods, after we apply balanced sampling for the source domain, the per-class average accuracy improved from 55.44\% to 58.50\%, which indicates balanced sampling is helpful for class distribution shift, despite only in the source domain.

\subsubsection{The effectiveness of implicit alignment}
The effectiveness of implicit alignment is demonstrated through the comparison between ``MDD+Implicit Alignment'' and ``MDD~(source-balanced sampler)''.
Both methods use the same sampling procedure for the source. The only difference is that implicit alignment aligns the two domains by sampling aligned classes in the target domain, whereas ``source-balanced sampler'' only takes random samples from the target domain.
Table~\ref{table:officehome-imbalanced} shows that implicit alignment performs better than ``source-balanced sampler'' because it is better-aligned, which confirms the effectiveness of implicit alignment.
Besides, the proposed method also outperforms MDD-based explicit alignment, which validates the effectiveness of implicit alignment over the explicit alignment.

\begin{table*}[htbp]
	%\addtolength{\tabcolsep}{2pt}
	\centering
	%\vspace{-5pt}
	\caption{Accuracy (\%) on {Office-31}~(standard) for unsupervised domain adaptation (ResNet-50). We repeated each experiment
5 times with different random seeds and report the average and the standard error of the
accuracy.}
	\label{table:office31}
	%\vskip 0.05in
	\scalebox{0.83}{%
	\begin{threeparttable}
		\begin{tabular}{lccccccc}
			\toprule
			Method                          & A $\shortrightarrow$ W     & D $\shortrightarrow$ W     & W $\shortrightarrow$ D     & A $\shortrightarrow$ D     & D $\shortrightarrow$ A     & W $\shortrightarrow$ A     & Avg           \\
			\midrule
			Source only  & 68.4$\pm$0.2          & 96.7$\pm$0.1          & 99.3$\pm$0.1          & 68.9$\pm$0.2          & 62.5$\pm$0.3          & 60.7$\pm$0.3          & 76.1          \\\midrule
			DAN \cite{long2015learning}       & 80.5$\pm$0.4          & 97.1$\pm$0.2          & 99.6$\pm$0.1          & 78.6$\pm$0.2          & 63.6$\pm$0.3          & 62.8$\pm$0.2          & 80.4          \\
			DANN~\cite{ganin2016domain}  & 82.0$\pm$0.4          & 96.9$\pm$0.2          & 99.1$\pm$0.1          & 79.7$\pm$0.4          & 68.2$\pm$0.4          & 67.4$\pm$0.5          & 82.2          \\
			ADDA \cite{tzeng2017adversarial}     & 86.2$\pm$0.5          & 96.2$\pm$0.3          & 98.4$\pm$0.3          & 77.8$\pm$0.3          & 69.5$\pm$0.4          & 68.9$\pm$0.5          & 82.9          \\
			JAN~\cite{long2017deep}       & 85.4$\pm$0.3          & {97.4}$\pm$0.2        & {99.8}$\pm$0.2        & 84.7$\pm$0.3          & 68.6$\pm$0.3          & 70.0$\pm$0.4          & 84.3          \\
			MADA~\cite{pei2018multi} &90.0 $\pm$ 0.1 & 97.4$\pm$0.1 & 99.6$\pm$0.1 & 87.8$\pm$0.2 & 70.3$\pm$0.3 & 66.4$\pm$0.3 & 85.2 \\ 
			GTA \cite{sankaranarayanan2018generate}       & 89.5$\pm$0.5          & 97.9$\pm$0.3          & 99.8$\pm$0.4          & 87.7$\pm$0.5          & 72.8$\pm$0.3          & 71.4$\pm$0.4          & 86.5          \\
			MCD~\cite{saito2018maximum}   &88.6$\pm$0.2&98.5$\pm$0.1&\textbf{100.0}$\pm$.0&92.2$\pm$0.2&69.5$\pm$0.1&69.7$\pm$0.3&86.5 \\
			CDAN \cite{long2018conditional}     & 94.1$\pm$0.1          & 98.6$\pm$0.1 & \textbf{100.0}$\pm$.0 & 92.9$\pm$0.2          & 71.0$\pm$0.3          & 69.3$\pm$0.3          & 87.7    \\
			MDD~\cite{pmlr-v97-zhang19i}                    & \textbf{94.5}$\pm$0.3 & 98.4$\pm$0.1          & \textbf{100.0}$\pm$.0 & \textbf{93.5}$\pm$0.2 & 74.6$\pm$0.3 & 72.2$\pm$0.1 & \textbf{88.9} \\
			PACET~\cite{liang2019exploring}$^{\color{red}{\ddagger}}$  & 90.8 & 97.6 & 99.8 & 90.8 & 73.5 & 73.6 & 87.4    \\
			CAT~\cite{deng2019cluster}$^{\color{red}{\ddagger}}$  & 94.4$\pm$0.1 & 98.0$\pm$0.2 & \textbf{100.0}$\pm$0.0 & 90.8$\pm$1.8 & 72.2$\pm$0.2 & 70.2$\pm$0.1 & 87.6     \\
			MDD (source-balanced sampler) & 90.4$\pm$0.4 & \textbf{98.7}$\pm$0.1 & 99.9$\pm$0.1  & 90.4$\pm$0.2 & 75.0$\pm$0.5& $73.7\pm$0.9 & 88.0 \\
			MDD+Explicit Alignment$^{\color{red}{\ddagger}}$ & 92.3$\pm$0.1 & 98.2$\pm$0.1 & 99.8$\pm$.0 & 92.3$\pm$0.3 & 74.6$\pm$0.2 & 72.9$\pm$0.7 & 88.4 \\
			\textbf{MDD+Implicit Alignment}                   & 90.3$\pm$0.2 & \textbf{98.7}$\pm$0.1          & 99.8$\pm$.0 & 92.1$\pm$0.5 & \textbf{75.3}$\pm$0.2 & \textbf{74.9}$\pm$0.3 & 88.8 \\
			\bottomrule
		\end{tabular}
		\begin{tablenotes}
    \item[$\color{red}{\ddagger}$] Methods using explicit class-conditioned domain alignment.
    
  \end{tablenotes}
		\end{threeparttable}
	}
	\vspace{-10pt}
\end{table*}
\addtolength{\tabcolsep}{-4pt}  
\begin{table*}[htbp]
	%\addtolength{\tabcolsep}{-5pt}
	\centering
	\caption{Accuracy (\%) on Office-Home~(standard) for unsupervised domain adaptation (ResNet-50).
	%If we compare the two methods that employ explicit alignment, ``MDD+Explicit Alignment'' outperforms MCS in 7 out of 12 source-target domain pairs. This indicates although MDD is a stronger adversarial training strategy, it is not able to benefit from explicit alignment. Same holds if we compare ``MDD'' with ``MDD+Explicit Alignment''.
	}
	\label{table:officehome}
%	\vskip 0.05in
	\scalebox{0.83}{%
	\begin{threeparttable}
		\begin{tabular}{lccccccccccccc}
			\toprule
			Method                          & Ar$\shortrightarrow$Cl & Ar$\shortrightarrow$Pr & Ar$\shortrightarrow$Rw & Cl$\shortrightarrow$Ar & Cl$\shortrightarrow$Pr & Cl$\shortrightarrow$Rw & Pr$\shortrightarrow$Ar & Pr$\shortrightarrow$Cl & Pr$\shortrightarrow$Rw & Rw$\shortrightarrow$Ar & Rw$\shortrightarrow$Cl & Rw$\shortrightarrow$Pr & Avg           \\
			\midrule
			Source only & 34.9                   & 50.0                   & 58.0                   & 37.4                   & 41.9                   & 46.2                   & 38.5                   & 31.2                   & 60.4                   & 53.9                   & 41.2                   & 59.9                   & 46.1          \\\midrule
			DAN~\cite{long2015learning}       & 43.6                   & 57.0                   & 67.9                   & 45.8                   & 56.5                   & 60.4                   & 44.0                   & 43.6                   & 67.7                   & 63.1                   & 51.5                   & 74.3                   & 56.3          \\
			DANN~\cite{ganin2016domain}  & 45.6                   & 59.3                   & 70.1                   & 47.0                   & 58.5                   & 60.9                   & 46.1                   & 43.7                   & 68.5                   & 63.2                   & 51.8                   & 76.8                   & 57.6          \\
			JAN~\cite{long2017deep}       & 45.9                   & 61.2                   & 68.9                   & 50.4                   & 59.7                   & 61.0                   & 45.8                   & 43.4                   & 70.3                   & 63.9                   & 52.4                   & 76.8                   & 58.3          \\
			CDAN~\cite{long2018conditional}     & 50.7                   & 70.6                   & 76.0                   & 57.6                   & 70.0                   & 70.0                   & 57.4                   & 50.9                   & 77.3                   & 70.9                   & 56.7                   & 81.6                   & 65.8   \\
			BSP~\cite{chen2019transferability} & 52.0 & 68.6 & 76.1 & 58.0 & 70.3 & 70.2 & 58.6 & 50.2 & 77.6 & 72.2 & 59.3 & 81.9 & 66.3 \\
			MDD~\cite{pmlr-v97-zhang19i}                & 54.9          & 73.7          & 77.8          & 60.0          &71.4          & 71.8          & 61.2          & 53.6          & 78.1          & \textbf{72.5}          & \textbf{60.2}          & 82.3          & 68.1
			\\
			MCS~\cite{liang2019distant}$^{\color{red}{\ddagger}}$ & 55.9 & 73.8 & 79.0 & 57.5 & 69.9 & 71.3 & 58.4 & 50.3 & 78.2 & 65.9 & 53.2 & 82.2 & 66.3
			\\
			MDD+Explicit Alignment$^{\color{red}{\ddagger}}$    &     54.3      &       74.6     &       77.6     &    60.7        &     71.9      &     71.4       &      62.1      &     52.4       &     76.9       &    71.1        &   57.6         &   81.3         & 67.7
			\\
			MDD (source-balanced sampler) & 55.3 & 75.0 & 79.1 & 62.3 & 70.1 & 73.2 & 63.5 & 53.2 & 78.7 & 70.4 & 56.2 & 82.0 & 68.3 \\
			\textbf{MDD+Implicit Alignment}                & \textbf{56.2}          & \textbf{77.9}          & \textbf{79.2}          & \textbf{64.4}          & \textbf{73.1}          & \textbf{74.4}          & \textbf{64.2}          & \textbf{54.2}          & \textbf{79.9}          &       71.2  & 58.1           & \textbf{83.1}          & \textbf{69.5}
			\\
			\bottomrule
		\end{tabular}%
		\begin{tablenotes}
    \item[$\color{red}{\ddagger}$] Methods using explicit class-conditioned domain alignment.
  \end{tablenotes}
		\end{threeparttable}
	}
	%\vspace{-5pt}
\end{table*}
\addtolength{\tabcolsep}{4pt}
%\textbf{Robustness to class imbalance.}
Figure~\ref{fig:imbalance_accuracy_gap}~(b) compares the baseline, implicit and explicit alignments on Office-Home~(balanced) and Office-Home~(RS-UT).
We observe that implicit alignment performs the best on both datasets.
%This table aims to show how basic MDD suffers from imbalance, and  implicit alignment is more robust to label imbalance, hence  a smaller gap.
More importantly, implicit alignment is more robust to class distribution shift which greatly out-performs other methods under RS-UT distribution shift and has a smaller performance drop from the balanced version of Office-Home.

\subsection{Evaluating Standard Domain Adaptation Datasets}
Table~\ref{table:office31} and Table~\ref{table:officehome} summarize the results for the standard Office-31 and Office-Home datasets which have a small degree of class imbalance.
Our method outperforms the baselines in 3 out of 6 domain pairs for Office-31, and 10 out of 12 domain pairs for Office-Home~(standard).
The proposed implicit alignment exhibits larger performance gains on the Office-Home dataset because the dataset is more difficult for domain adaptation, and it has 65 classes compared with the 31 classes in Office-31.
%Our method is especially useful for tasks with a large number of classes because not all classes can fit in one batch, which necessitates sampling-based alignment for better training.
We also report state-of-the-art results for VisDA in Table~\ref{tab:visda}.

%\subsubsection{The impact of source-balanced sampling}
Similar to findings in~Section~\ref{sec:extreme-class-distribution-shift}, we observe source-balanced sampling is helpful when comparing ``MDD (source-balanced sampler)'' with the MDD standard baseline, even without extreme class distribution shift.

%\subsubsection{Comparison with explicit alignment}
The proposed method outperforms the state-of-the-art explicit alignment methods---PACET and MCS---across all domain pairs.
We find it ineffective to incorporate prototype-based explicit alignment into MDD.
This is in contrast with domain-discriminator-based adversarial learning, where explicit alignment is shown to improve domain adaptation.
This is because the classifier-based discrepancy MDD contains more abundant information than domain-discriminator-based discrepancy, owing to the availability of predictive probabilities provided by the classifiers.
The rich information in domain discrepancy removes the need for prototype-based distances.
\begin{table}[t]
\centering
\renewcommand{\arraystretch}{1.2}
    \scalebox{0.9}{
\begin{tabular}{cc}
\toprule
method & acc.~(\%) \\\midrule
JAN~\cite{long2017deep} & 61.6 \\
GTA\cite{sankaranarayanan2018generate} & 69.5 \\
MCD~\cite{saito2018maximum} & 69.8 \\
CDAN~\cite{long2018conditional} & 70.0 \\
MDD~\cite{pmlr-v97-zhang19i} & 74.6 \\
%MDD~(re-implementation) & 68.7 \\
MDD+Explicit Alignment & 67.1 \\
\textbf{MDD+Implicit Alignment} & \textbf{75.8} \\\bottomrule
\end{tabular}
}
\caption{VisDA2017 target accuracy (ResNet-50)}
\label{tab:visda}
\end{table}
%Moreover, we find that explicit alignment is very sensitive to the weight $\lambda$ of the alignment loss. We experimented with different values of  $\lambda\in[0.1, 0.01, 0.001,0.0001,0.00001]$ and chose $0.0001$ as the best performing one for ``MDD+Explicit Alignment''.

%We provide more evaluation measures in the supplementary materials.
%and show that the improvements from our model are not a result of batch size or random seeds.
%under identical batch sizes and random seeds.
%The results confirm that the improvements from our model are not a result of batch size or random seeds.

\begin{figure*}[ht]
\begin{minipage}[!t]{0.3\linewidth}
\vspace{0pt}
    %\centering
    \includegraphics[width=0.95\textwidth]{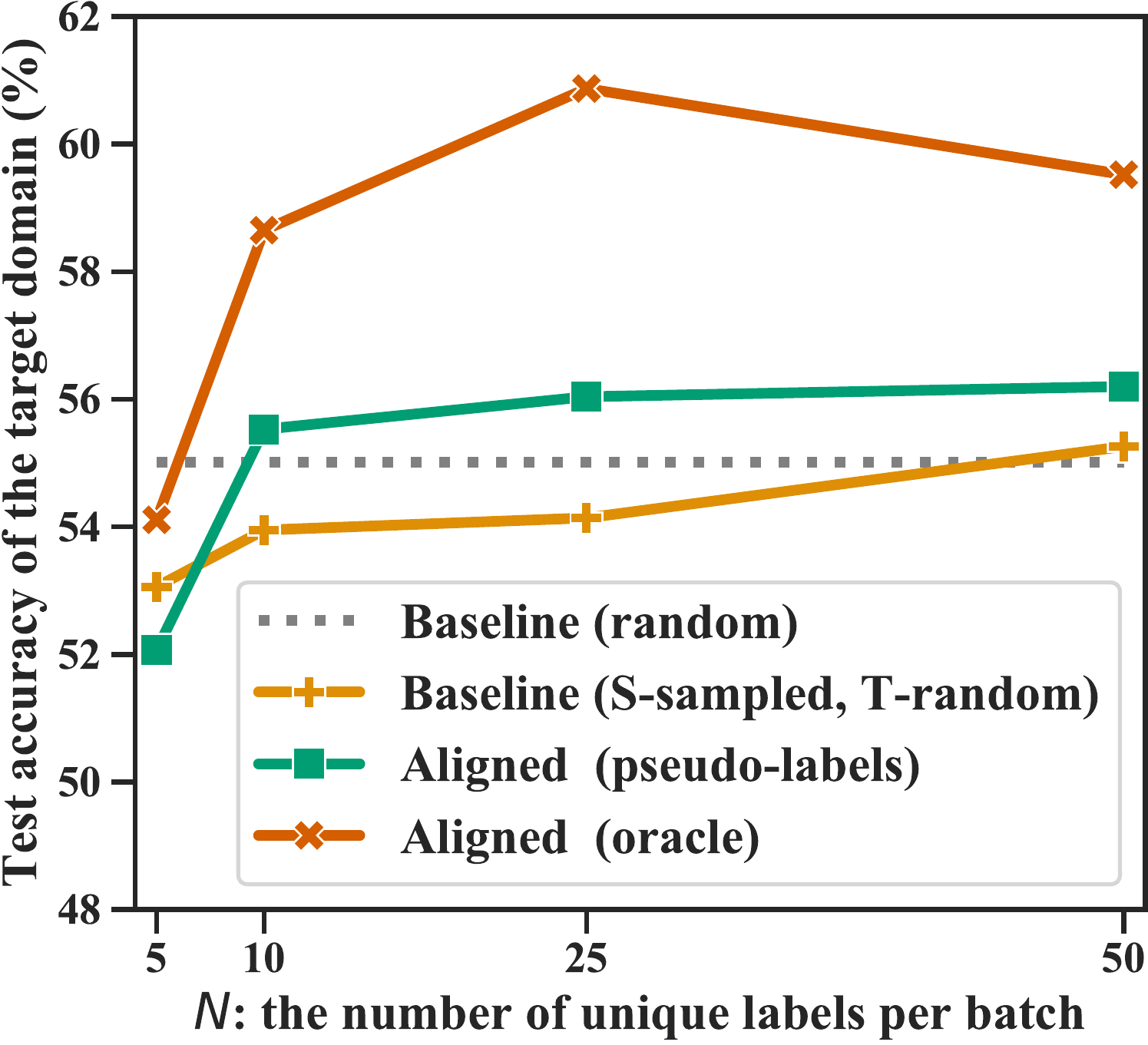}
    \captionof{figure}{The impact of class diversity and alignment on domain adaptation for Ar$\rightarrow$Cl, Office-Home~(standard).}
    \label{fig:impact_of_cls_diversity}
\end{minipage}\hfill
%\hspace{0.5cm}
\begin{minipage}[!t]{0.3\linewidth}
\vspace{0pt}
    %\centering
    \includegraphics[width=0.95\textwidth]{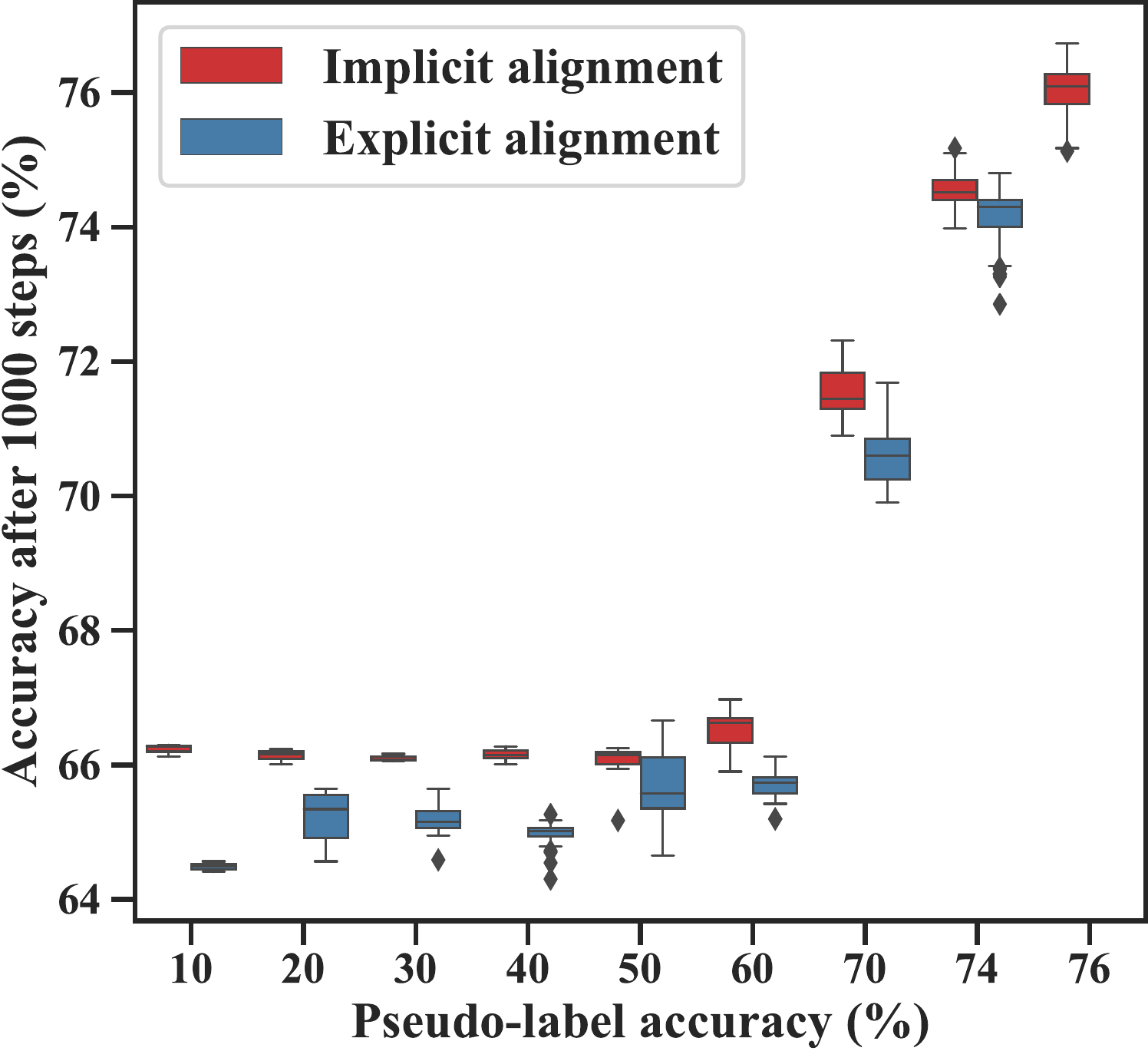}
    \captionof{figure}{
    The impact of pseudo-label errors on implicit and explicit alignment,  Ar$\rightarrow$Cl, Office-Home~(standard).
    %Implicit alignment is more robust to pseudo-label errors  than explicit alignment for Ar$\rightarrow$Pr,~Office-Home~(standard).
    }
    \label{fig:robustness_to_calibration}
\end{minipage}\hfill
\begin{minipage}[!t]{0.32\linewidth}
\vspace{0pt}\raggedright
\centering
\scalebox{0.8}{
\begin{tabular}[!t]{cccc}
\toprule
 & \multicolumn{2}{c}{Alignment options} & \\ \cmidrule{2-3}
Domains                                 & masking               & sampling              & avg. acc. \\\midrule
\multirow{4}{*}{Rw$\shortrightarrow$Cl} & $\times$   & $\times$    & 44.8                 \\
                                        & $\surd$   & $\times$    & 44.8                 \\
                                        & $\times$   & $\surd$    & 47.4                 \\
                                        & $\surd$   & $\surd$    & \textbf{50.0}                 \\\midrule
\multirow{4}{*}{Pr$\shortrightarrow$Rw} & $\times$   & $\times$    & 69.3                 \\
                                        & $\surd$   & $\times$    & 72.7                 \\
                                        & $\times$   & $\surd$    & 72.0                 \\
                                        & $\surd$   & $\surd$    & \textbf{74.2}       \\\bottomrule         

\end{tabular}
}
\captionof{table}{The impact of different implicit alignment options, i.e., masking in the MDD estimation and sampling class-aligned minibatches, on Office-Home~(RS-UT). }
\label{tab:alignment-ablation}
\end{minipage}
\vspace{-10pt}
\end{figure*}
\subsection{Ablation studies}
\subsubsection{Impact of class diversity and alignment}
We analyze the impact of class diversity and alignment by designing experiments along three dimensions: the number of unique labels in each minibatch, whether the classes are aligned, and whether we use pseudo-labels or ground-truth labels when sampling the target domain.
%Note that all experiments in Figure~\ref{fig:impact_of_cls_diversity} have the same batch size 50.

\textbf{Setup.}
``Baseline~(random)'' randomly samples examples of both domains.
``Baseline~(S-sampled, T-random)'' uses $N$-way sampler for the source domain, and randomly samples the target domain.
``Aligned~(pseudo-labels)'' is the proposed implicit alignment approach.
% where the source domain is sampled with $N$-way $K$-shot samplers, and the target domain is sampled based on pseudo-labels from model predictions.
``Aligned~(Oracle)'' is the oracle form of implicit alignment where the target domain uses ground-truth labels for sampling.

\textbf{The impact of class diversity.}
Minibatch-based class diversity determines the sampling distribution of the label space, and a greater diversity corresponds to a more stable measure of this sampling distribution.
Figure~\ref{fig:impact_of_cls_diversity} suggests a positive correlation between the model performance and class diversity: domain adaptation methods do not work well when the class diversity is very low---i.e., only sample 5 classes per batch among the 65 classes---and the alignment-based methods outperform the baseline as we increase class diversity.

\textbf{The impact of alignment.}
We confirm the importance of the proposed implicit alignment algorithm from two perspectives.
First, ``Aligned (oracle)'' consistently performs the best, which suggests perfect alignment can provide substantial benefits to unsupervised domain adaptation.
Second, the comparison between ``Aligned~(pseudo-labels)'' and ``Baseline~(S-sampled, T-random)'' validates the effectiveness of pseudo-label based implicit alignment, although the pseudo-labels are approximations of the oracle.
%As noted in previous experiments, aligning the source and target domains is beneficial to domain adaptation.
%Moreover, we have consistent finding by comparing ``Baseline~(S-sampled, T-random)'' with ``Aligned~(pseudo-labels)'', given sufficient class diversity (i.e., $N>5$).

\subsubsection{Robustness to pseudo-label errors}
We investigate whether implicit alignment is indeed more robust to pseudo-label errors when compared with explicit alignment.
Figure~\ref{fig:robustness_to_calibration} illustrates the relationship between pseudo-label accuracy at training step $t$ and the corresponding subsequent target accuracy at step $t+1000$, i.e., after 1000 domain adaptation training steps.
This process resembles a Markov chain that allows us to analyze the impact of pseudo-label accuracy on the learning dynamics.
%through the target domain accuracy.%, given the same number of training steps.

It is evident that the drawbacks of explicit alignment are more severe when the pseudo-labels are less accurate, e.g., 10$\sim$40\%, where implicit alignment has more considerable performance improvements than explicit alignment.
This suggests that implicit alignment is more robust to erroneous pseudo-label predictions because it does not require explicit supervision from the pseudo-labels.
Implicit and explicit methods eventually converge at 76\% and 74\%, respectively.
%Implicit alignment eventually converges at 76\% while the explicit alignment converges at 74\% accuracy.

Although many recent techniques attempt to address pseudo-label bias in explicit alignment, they depend on the assumption that probabilities of model predictions are well-calibrated during training.
They fail to address ill-calibrated probabilities~\cite{guo2017calibration}, where the model tends to make confident mistakes on the target domain.
Moreover, given that models do not initially perform well when training begins, for a random classifier, implicit alignment selects random samples that is equivalent to training without sampling. In contrast, explicit alignment optimizes model parameters from these random labels explicitly.
%
%\emph{The impact of random classifiers.}
%Implicit alignment provides random samples from a dataset with random pseudo-labels, which does not introduce pseudo-label bias into model optimization.
%A random classifier assigns random examples to each class.
%Therefore, implicit alignment provides random samples of the dataset, which do not introduce pseudo-label bias into model optimization.
%Therefore, we can deploy implicit alignment once we start training the model, even from randomly initialized parameters.
%This is in contrast with explicit alignment that directly optimizes model parameters from pseudo-labels, which might lead to error accumulation.
\subsubsection{Ablation Study onn MDD}
Table~\ref{tab:alignment-ablation} presents the ablation study on Office-Home~(RS-UT) that aims to assess the impact of different implicit alignment options: alignment in the domain divergence estimations in Section~\ref{sec:minibatch-sampling-strategy}~(i.e., \emph{masking} in MDD) and alignment in the input space in Section~\ref{sec:minibatch-sampling-strategy}~(i.e., \emph{sampling} class-conditioned examples).
We observe that both alignment techniques are essential for domain adaptation because alignment should be enforced consistently across all aspects of adaptation.
We report similar findings, in the supplementary material, on Office-Home~(standard).

\begin{figure}
    \centering
    \includegraphics[width=0.35\textwidth]{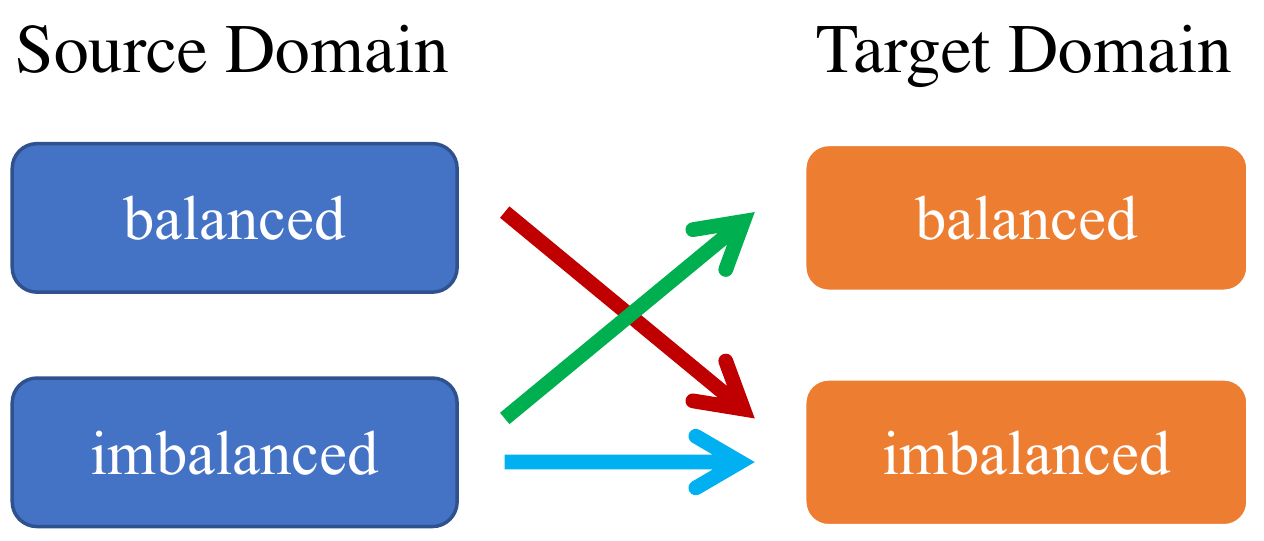}
    \caption{Interactions between \emph{within-domain class imbalance} and \emph{between-domain class distribution shift}.}
    \label{fig:imbalance_illustration}
\end{figure}
\subsubsection{Generalization: implicit alignment also improves DANN}
We design additional experiments to further demonstrate the effectiveness of the proposed approach on \emph{a different domain adaptation algorithm}---DANN---on two synthetic domains with \emph{different degrees of class imbalance}: ``mild''~(light-tailed class imbalance from a triangular-like distribution) and ``extreme''~(heavy-tailed class imbalance from a Pareto distribution).
We synthetically manipulate the class distributions of SVHN and MNIST to simulate various interactions between \emph{within-domain class imbalance} and \emph{between-domain class distribution shift}.
As illustrated in Fig~\ref{fig:imbalance_illustration}, we simulate three types of distribution shift when $p_S(y)\neq p_T(y)$ (i)~{\color{BrickRed}source-balanced, target-imbalanced}; (ii)~{\color{ForestGreen}source-imbalanced, target-balanced}; (iii)~{\color{Cyan}both-imbalanced}.
%We choose the domain pair MNIST and SVHN as they are the most challenging domain pairs in the digits domain adaptation dataset.
\begin{table}[t]
    \centering
\caption{Per-class average accuracy~(\%) with \emph{mismatched prior} where the source domain is balanced while the target domain is imbalanced.}
\vspace{4pt}
\addtolength{\tabcolsep}{-4pt} 
\scalebox{0.9}{
\begin{tabular}{cccccc}
\toprule
\multicolumn{1}{l}{} & \multicolumn{2}{c}{SVHN$\rightarrow$MNIST}                                                                                           &  & \multicolumn{2}{c}{MNIST$\rightarrow$SVHN}               \\ \cline{2-3} \cline{5-6}
method                                            & mild                        & extreme                        &  & mild & extreme \\ \midrule
source only & 67.4$\pm$7.3  & 66.3$\pm$3.3 & & \textbf{32.5}$\pm$2.9  & \textbf{28.2}$\pm$2.3 \\
DANN & 78.2$\pm$2.8  & 59.1$\pm$0.8 & & 20.9$\pm$6.0  & 20.5$\pm$3.1  \\
DANN+implicit & \textbf{88.6}$\pm$0.7 & \textbf{82.2}$\pm$2.1 &  & \textbf{32.4}$\pm$2.1 & \textbf{28.9}$\pm$3.3 \\
\bottomrule
\end{tabular}
}
\label{tab:digits-mismatchprior-sourceB-targetIm}
\end{table}

Table~\ref{tab:digits-mismatchprior-sourceB-targetIm},~\ref{tab:digits-mismatchprior-sourceIm-targetB}~and~\ref{tab:digits-mismatchprior-both-im} present the results for the abovementioned scenarios and all experiments are repeated five times.
The proposed implicit alignment approach significantly improves the performance of DANN regardless of the degree of imbalance or the type of distribution shift.
Besides, implicit alignment offers greater improvements over DANN when the degree of imbalance is more severe, i.e., comparing ``mild'' with ``extreme''.
Implicit alignment overcomes this limitation of DANN and greatly improves the performance of the challenging task between SVHN and MNIST.
We conclude that the proposed approach is independent of the choice of domain adaptation algorithms and helps both MDD and DANN.

Note that the aim of this subsection is to show that implicit alignment could help improve DANN on the digits dataset.
More work is needed to compare with the current state-of-the-art methods~\cite{kumar2018co,shu2018dirt} on this dataset.
\begin{table}[t]
\caption{Per-class average accuracy~(\%) with \emph{mismatched prior} where the source domain is imbalanced while the target domain is balanced.}
\vspace{4pt}
\addtolength{\tabcolsep}{-4pt} 
\renewcommand{\arraystretch}{1.2}
\centering
\scalebox{0.9}{
\begin{tabular}{cccccc}
\toprule
\multicolumn{1}{l}{} & \multicolumn{2}{c}{SVHN$\rightarrow$MNIST}                                                                                           &  & \multicolumn{2}{c}{MNIST$\rightarrow$SVHN}               \\ \cline{2-3} \cline{5-6}
method                                            & mild                        & extreme                        &  & mild & extreme \\ \midrule
source only  & 65.2$\pm$2.1  & 53.3$\pm$1.3  & & \textbf{31.6}$\pm$3.3  & \textbf{32.8}$\pm$0.9 \\
DANN & 82.0$\pm$0.7  & 52.3$\pm$2.3 & & 23.4$\pm$3.6  & 25.9$\pm$0.5 \\
DANN+implicit & \textbf{91.0}$\pm$1.9 & \textbf{87.1}$\pm$2.6 & & \textbf{34.9}$\pm$0.5 & \textbf{31.1}$\pm$2.9 \\
\bottomrule
\end{tabular}
}
\label{tab:digits-mismatchprior-sourceIm-targetB}
\end{table}
\begin{table}[t]
    \centering
\caption{Per-class average accuracy~(\%) with \emph{mismatched prior} where both domains are imbalanced.}
\vspace{4pt}
\scalebox{0.9}{
\addtolength{\tabcolsep}{-4pt} 
\begin{tabular}{cccccc}
\toprule
\multicolumn{1}{l}{} & \multicolumn{2}{c}{SVHN$\rightarrow$MNIST}                                                                                           &  & \multicolumn{2}{c}{MNIST$\rightarrow$SVHN}               \\ \cline{2-3} \cline{5-6}
method                                            & mild                        & extreme                        &  & mild & extreme \\ \midrule
source only   & 60.9$\pm$5.2 & 51.2$\pm$5.9 & & 30.6$\pm$1.3 & \textbf{27.1}$\pm$1.7 \\
DANN  & 67.6$\pm$0.8  & 40.5$\pm$5.5 & & 23.4$\pm$1.6  & 18.8$\pm$2.9  \\
DANN+implicit & \textbf{88.6}$\pm$0.6 & \textbf{70.5}$\pm$3.6 & & \textbf{36.3}$\pm$2.5 & \textbf{27.9}$\pm$2.4 \\
\bottomrule
\end{tabular}
}
\label{tab:digits-mismatchprior-both-im}
\end{table}

\section{Related Work}
We review related work on unsupervised domain adaptation and discuss their relations with our proposed method.

%{\color{blue}Talk about the theoretical basis.}

%\subsection{Domain Adaptation}

%\emph{Domain translation}~\cite{liu2017unsupervised} is a type of input-based domain adaptation method that aims at translating images from the source domain to images in the target domain, which is often constrained by cycle-consistencies~\cite{hoffman2017cycada}.
%During inference, a target example is first translated to the source domain, then predicted with the classifier trained on the source domain.

\emph{Instance-based importance-weighting}~\cite{chawla2002smote,kouw2019review} aims to minimize the target error directly from the source domain data, weighted at the example level or class level.
%Example-level weighting is designed to address covariate shift that uses important-weighting to train the classifier from the source domain data.
%It first estimates of the probability of a source example belonging to the target domain, then uses that probability as important-weighting to train the classifier from the source domain data.
%Conversely, class-level weighting applies the weighting on class labels and is designed to address cost-sensitive learning~\cite{elkan2001foundations} and class imbalance.
%Oversampling methods~\cite{chawla2002smote} have an equivalent effect with importance-weighting.
%Our proposed method is more in line with sampling-based importance-weighting.
Unlike our approach, importance-weighting only uses the source data to train the classifier without learning domain invariant representations.

\emph{Feature-based distribution adaptation} is the prevailing approach to domain adaptation that aims to minimize the distribution discrepancy between the source and target domains.
The domain difference can be measured in various ways, such as Maximum Mean Discrepancy~(MMD)~\cite{borgwardt2006integrating}, which is further minimized to achieve domain invariance.
%and Wasserstein distance~\cite{shen2018wasserstein}, which is further minimized to achieve domain invariance.
The minimization of such discrepancy can be carried out by directly minimizing the distance~\cite{tzeng2014deep} or with the help of adversarial learning~\cite{ganin2016domain}.

\emph{Classifier-based distribution adaptation} is a strong competitor to feature-based adaptation.
It aims to minimize the discrepancy between two classifiers so that the learned representations respect the decision boundary of the classification task~\cite{saito2018maximum,pmlr-v97-zhang19i}.
%An alternative classifier-based approach is to minimize the conditional entropy of model predictions by assuming a prior that favors minimal class overlap on the unlabeled data~\cite{grandvalet2005semi,luo2017label}.
%We use classifier-based discrepancy MDD for adversarial training because the probabilities predicted by two classifiers are more informative than domain discriminator-based binary outcomes.
We show that the proposed approach is beneficial to both classifier-based discrepancy MDD~\cite{pmlr-v97-zhang19i} and feature-based discrepancy DANN~\cite{ganin2016domain}.

\emph{Feature-classifier joint distribution adaptation} aims to align the joint distribution between features and their corresponding predictions~\cite{long2013transfer,tsai2018learning}.
The joint distribution can be represented in a multilinear map between features and classifier predictions~\cite{long2018conditional}, or the Cartesian product between the domain space and class space~\cite{Cicek_2019_ICCV}.
In our work, we implicitly align the joint distribution with the factorization $p(x,y)=p(x|y)p(y)$ from a sampling perspective where $p(y)$ is the pre-specified alignment distribution in the label space, and $p(x|y)$ represents class-conditioned sampling.
%\cite{Cicek_2019_ICCV} proposes to learn a joint distribution between the domain label and class label,  and aims to create class confusion in the Cartesian product between the domain space and class space.
%Conditional Domain Adversarial Adaptation Network~(CDAN) proposes to use a conditional domain discriminator on the multilinear maps of features and classifier predictions~\cite{long2018conditional}.
%Joint distribution adaptation~\cite{long2013transfer} aims to adapt both the conditional predictive distribution~$P(Y|X)$ and marginal distribution of the data~$P(X)$ using MMD, where pseudo-labels are used to approximate~$Q(Y^t|X^t)$.
%But in their implementation, they use class-conditional distribution $Q(X^t|Y^t)$ instead of the posterior $Q(Y^t|X^t)$.

\emph{Explicit class-conditioned domain alignment}, or class prototype alignment, introduces a loss function that minimizes the distances of class-level prototypes between the source and target domains~\cite{snell2017prototypical,pinheiro2018unsupervised,pan2019transferrable,deng2019cluster}.
It is prone to error accumulation due to its reliance on explicit optimization of model parameters from the pseudo-labels.
A variety of recent methods have been proposed to mitigate these limitations by estimating batch-level statistics~\cite{xie2018learning} and introducing an easy-to-hard curriculum that favors confident predictions~\cite{chen2019progressive}.
Nevertheless, these algorithms suffer from ill-calibrated probabilities in the form of confident mistakes, and more work is needed to improve model calibration so as to better utilize explicit alignment.

\emph{Self-training}~\cite{nigam2000analyzing} is a special form of co-training~\cite{blum1998combining} where the model iteratively uses its predictions, i.e., pseudo-labels, as explicit supervision to re-train itself.
The use of pseudo-labels has become an emerging trend in domain adaptation, because they provide estimations of the target domain label distribution that can be exploited by training algorithms.
Apart from class prototype based methods~\cite{chen2011co,saito2017asymmetric,zhang2018collaborative,deng2019cluster} for explicit alignment, \cite{wen2019bayesian} proposed the use of uncertainty estimates of the target domain predictions as second-order statistics to promote feature-label joint adaptation.
For semantic segmentation tasks, \citep{zou2018unsupervised} proposed to iteratively generate pseudo-labels in the target domain and re-train the model on these labels; \citep{zhang2019category} proposed to use pseudo-labels to encourage examples to cluster together if they belong to the same class; \citep{chen2019domain} applied entropy minimization~\cite{grandvalet2005semi} on the pseudo-labels to encourage class overlap between domains.
A main bottleneck for this approach is the bias in pseudo-label predictions. Directly optimizing these labels is prone to ``entropy over-minimization''~\cite{Zou_2019_ICCV} and negative transfer~\cite{lifshitz2020sample} where the model overfits to mistakes in the pseudo-labels.
%, which requires confidence regularization~\cite{Zou_2019_ICCV} and learning curriculum of the confidence threshold.
% and an easy- curriculum~\cite{chen2019progressive} to facilitate learning.
Moreover, the pseudo-labels are likely to suffer from ill-calibrated probabilities~\cite{guo2017calibration}, especially for deep learning methods.
The resulting misleadingly confident mistakes exacerbate the critical problem of error accumulation in pseudo-label bias.
In contrast, our proposed method removes the need for direct supervision from pseudo-labels, and as a result is more robust to bias in how these labels are produced.

\emph{Reinforced sample selection}~\cite{dong2018domain} is proposed for one-shot domain adaptation where a model actively selects labeled examples to train the domain adaptation model.
In comparison, the advantage of our approach is in its simplicity that no reinforcement learning is required to obtain the sampling strategy.
\section{Conclusion and Future Work}
We introduce an approach for unsupervised domain adaptation---with a strong focus on practical considerations of within-domain class imbalance and between-domain class distribution shift---from a class-conditioned domain alignment perspective.
We show theoretically that the proposed implicit alignment provides a more reliable measure of empirical domain divergence which facilitates adversarial domain-invariant representation learning, that would otherwise be hampered by the class-misaligned domain divergence.
%We extend our theory on implicit alignment into the classifier-based domain divergence measure and provide extensive experiments to 
We show that our proposed approach leads to superior UDA performance under extreme within-domain class imbalance and between-domain class distribution shift, as well as competitive results on standard UDA tasks.
%We further demonstrate that implicit alignment overcomes the critical limitations of pseudo-label bias by removing the need for explicit optimization of model parameters from pseudo-labels.
We emphasize that the proposed method is robust to pseudo-label bias, simple to implement, has a unified training objective, and does not require additional parameter tuning.
We also show that the proposed approach is orthogonal to the choice of domain adaptation algorithms and offers consistent improvements to feature-based and classifier-based domain adaptation algorithms.

%Future work includes extensions to other domain adaptation setups where the label spaces between the source and target domains are not identical.
Future work includes extensions to cost-sensitive learning for domain adaptation, and other setups where the label space between the source and target domains are not identical, as well as other domain adaptation setups~\cite{cao2018partial}.
%where the label space of the two domains are not identical.
%apply implicit alignment to other domain adaptation setups, such as universal domain adaptation where the source and target domains have a common label space as well as a private domain-specific label space~\cite{you2019universal}.
It is also important to analyze the probability calibration of different domain adaptation models and develop well-calibrated methods for more effective use of pseudo-labels.

\section*{Acknowledgements}
We thank the anonymous reviewers for providing thoughtful feedback. The authors also thank Lisa Di Jorio, Tanya Nair, Francis Dutil, Cecil Low-Kam, Nicolas Chapados, and the Imagia team for their support. Xiang Jiang acknowledges the support of NVIDIA Corporation with the donation of the Titan X GPU used for this research.

\bibliography{ref}

\begin{thebibliography}{58}
\providecommand{\natexlab}[1]{#1}
\providecommand{\url}[1]{\texttt{#1}}
\expandafter\ifx\csname urlstyle\endcsname\relax
  \providecommand{\doi}[1]{doi: #1}\else
  \providecommand{\doi}{doi: \begingroup \urlstyle{rm}\Url}\fi

\bibitem[Ben-David et~al.(2010)Ben-David, Blitzer, Crammer, Kulesza, Pereira,
  and Vaughan]{ben2010theory}
Ben-David, S., Blitzer, J., Crammer, K., Kulesza, A., Pereira, F., and Vaughan,
  J.~W.
\newblock A theory of learning from different domains.
\newblock \emph{Machine learning}, 79\penalty0 (1-2):\penalty0 151--175, 2010.

\bibitem[Blum \& Mitchell(1998)Blum and Mitchell]{blum1998combining}
Blum, A. and Mitchell, T.
\newblock Combining labeled and unlabeled data with co-training.
\newblock In \emph{Proceedings of the eleventh annual conference on
  Computational learning theory}, pp.\  92--100. ACM, 1998.

\bibitem[Borgwardt et~al.(2006)Borgwardt, Gretton, Rasch, Kriegel,
  Sch{\"o}lkopf, and Smola]{borgwardt2006integrating}
Borgwardt, K.~M., Gretton, A., Rasch, M.~J., Kriegel, H.-P., Sch{\"o}lkopf, B.,
  and Smola, A.~J.
\newblock Integrating structured biological data by kernel maximum mean
  discrepancy.
\newblock \emph{Bioinformatics}, 22\penalty0 (14):\penalty0 e49--e57, 2006.

\bibitem[Cao et~al.(2018)Cao, Ma, Long, and Wang]{cao2018partial}
Cao, Z., Ma, L., Long, M., and Wang, J.
\newblock Partial adversarial domain adaptation.
\newblock In \emph{Proceedings of the European Conference on Computer Vision
  (ECCV)}, pp.\  135--150, 2018.

\bibitem[Chawla et~al.(2002)Chawla, Bowyer, Hall, and
  Kegelmeyer]{chawla2002smote}
Chawla, N.~V., Bowyer, K.~W., Hall, L.~O., and Kegelmeyer, W.~P.
\newblock Smote: synthetic minority over-sampling technique.
\newblock \emph{Journal of artificial intelligence research}, 16:\penalty0
  321--357, 2002.

\bibitem[Chen et~al.(2019{\natexlab{a}})Chen, Xie, Huang, Rong, Ding, Huang,
  Xu, and Huang]{chen2019progressive}
Chen, C., Xie, W., Huang, W., Rong, Y., Ding, X., Huang, Y., Xu, T., and Huang,
  J.
\newblock Progressive feature alignment for unsupervised domain adaptation.
\newblock In \emph{Proceedings of the IEEE Conference on Computer Vision and
  Pattern Recognition}, pp.\  627--636, 2019{\natexlab{a}}.

\bibitem[Chen et~al.(2011)Chen, Weinberger, and Blitzer]{chen2011co}
Chen, M., Weinberger, K.~Q., and Blitzer, J.
\newblock Co-training for domain adaptation.
\newblock In \emph{Advances in neural information processing systems}, pp.\
  2456--2464, 2011.

\bibitem[Chen et~al.(2019{\natexlab{b}})Chen, Xue, and Cai]{chen2019domain}
Chen, M., Xue, H., and Cai, D.
\newblock Domain adaptation for semantic segmentation with maximum squares
  loss.
\newblock In \emph{Proceedings of the IEEE International Conference on Computer
  Vision}, pp.\  2090--2099, 2019{\natexlab{b}}.

\bibitem[Chen et~al.(2019{\natexlab{c}})Chen, Wang, Long, and
  Wang]{chen2019transferability}
Chen, X., Wang, S., Long, M., and Wang, J.
\newblock Transferability vs. discriminability: Batch spectral penalization for
  adversarial domain adaptation.
\newblock In \emph{International Conference on Machine Learning}, pp.\
  1081--1090, 2019{\natexlab{c}}.

\bibitem[Cicek \& Soatto(2019)Cicek and Soatto]{Cicek_2019_ICCV}
Cicek, S. and Soatto, S.
\newblock Unsupervised domain adaptation via regularized conditional alignment.
\newblock In \emph{The IEEE International Conference on Computer Vision
  (ICCV)}, October 2019.

\bibitem[Deng et~al.(2019)Deng, Luo, and Zhu]{deng2019cluster}
Deng, Z., Luo, Y., and Zhu, J.
\newblock Cluster alignment with a teacher for unsupervised domain adaptation.
\newblock In \emph{Proceedings of the IEEE International Conference on Computer
  Vision}, pp.\  9944--9953, 2019.

\bibitem[Dong \& Xing(2018)Dong and Xing]{dong2018domain}
Dong, N. and Xing, E.~P.
\newblock Domain adaption in one-shot learning.
\newblock In \emph{Joint European Conference on Machine Learning and Knowledge
  Discovery in Databases}, pp.\  573--588. Springer, 2018.

\bibitem[Ganin et~al.(2016)Ganin, Ustinova, Ajakan, Germain, Larochelle,
  Laviolette, Marchand, and Lempitsky]{ganin2016domain}
Ganin, Y., Ustinova, E., Ajakan, H., Germain, P., Larochelle, H., Laviolette,
  F., Marchand, M., and Lempitsky, V.
\newblock Domain-adversarial training of neural networks.
\newblock \emph{The Journal of Machine Learning Research}, 17\penalty0
  (1):\penalty0 2096--2030, 2016.

\bibitem[Grandvalet \& Bengio(2005)Grandvalet and Bengio]{grandvalet2005semi}
Grandvalet, Y. and Bengio, Y.
\newblock Semi-supervised learning by entropy minimization.
\newblock In \emph{Advances in neural information processing systems}, pp.\
  529--536, 2005.

\bibitem[Guo et~al.(2017)Guo, Pleiss, Sun, and Weinberger]{guo2017calibration}
Guo, C., Pleiss, G., Sun, Y., and Weinberger, K.~Q.
\newblock On calibration of modern neural networks.
\newblock In \emph{Proceedings of the 34th International Conference on Machine
  Learning-Volume 70}, pp.\  1321--1330. JMLR. org, 2017.

\bibitem[He et~al.(2016)He, Zhang, Ren, and Sun]{he2016deep}
He, K., Zhang, X., Ren, S., and Sun, J.
\newblock Deep residual learning for image recognition.
\newblock In \emph{Proceedings of the IEEE conference on computer vision and
  pattern recognition}, pp.\  770--778, 2016.

\bibitem[Heckman(1979)]{heckman1979sample}
Heckman, J.~J.
\newblock Sample selection bias as a specification error.
\newblock \emph{Econometrica: Journal of the econometric society}, pp.\
  153--161, 1979.

\bibitem[Kouw \& Loog(2019)Kouw and Loog]{kouw2019review}
Kouw, W.~M. and Loog, M.
\newblock A review of domain adaptation without target labels.
\newblock \emph{IEEE transactions on pattern analysis and machine
  intelligence}, 2019.

\bibitem[Kumar et~al.(2018)Kumar, Sattigeri, Wadhawan, Karlinsky, Feris,
  Freeman, and Wornell]{kumar2018co}
Kumar, A., Sattigeri, P., Wadhawan, K., Karlinsky, L., Feris, R., Freeman, B.,
  and Wornell, G.
\newblock Co-regularized alignment for unsupervised domain adaptation.
\newblock In \emph{Advances in Neural Information Processing Systems}, pp.\
  9345--9356, 2018.

\bibitem[Liang et~al.(2019{\natexlab{a}})Liang, He, Sun, and
  Tan]{liang2019distant}
Liang, J., He, R., Sun, Z., and Tan, T.
\newblock Distant supervised centroid shift: A simple and efficient approach to
  visual domain adaptation.
\newblock In \emph{Proceedings of the IEEE Conference on Computer Vision and
  Pattern Recognition}, pp.\  2975--2984, 2019{\natexlab{a}}.

\bibitem[Liang et~al.(2019{\natexlab{b}})Liang, He, Sun, and
  Tan]{liang2019exploring}
Liang, J., He, R., Sun, Z., and Tan, T.
\newblock Exploring uncertainty in pseudo-label guided unsupervised domain
  adaptation.
\newblock \emph{Pattern Recognition}, 96:\penalty0 106996, 2019{\natexlab{b}}.

\bibitem[Lifshitz \& Wolf(2020)Lifshitz and Wolf]{lifshitz2020sample}
Lifshitz, O. and Wolf, L.
\newblock A sample selection approach for universal domain adaptation.
\newblock \emph{arXiv preprint arXiv:2001.05071}, 2020.

\bibitem[Lipton et~al.(2018)Lipton, Wang, and Smola]{lipton2018detecting}
Lipton, Z.~C., Wang, Y.-X., and Smola, A.
\newblock Detecting and correcting for label shift with black box predictors.
\newblock \emph{arXiv preprint arXiv:1802.03916}, 2018.

\bibitem[Long et~al.(2013)Long, Wang, Ding, Sun, and Yu]{long2013transfer}
Long, M., Wang, J., Ding, G., Sun, J., and Yu, P.~S.
\newblock Transfer feature learning with joint distribution adaptation.
\newblock In \emph{Proceedings of the IEEE international conference on computer
  vision}, pp.\  2200--2207, 2013.

\bibitem[Long et~al.(2015)Long, Cao, Wang, and Jordan]{long2015learning}
Long, M., Cao, Y., Wang, J., and Jordan, M.~I.
\newblock Learning transferable features with deep adaptation networks.
\newblock In \emph{Proceedings of the 32nd International Conference on
  International Conference on Machine Learning-Volume 37}, pp.\  97--105. JMLR.
  org, 2015.

\bibitem[Long et~al.(2017)Long, Zhu, Wang, and Jordan]{long2017deep}
Long, M., Zhu, H., Wang, J., and Jordan, M.~I.
\newblock Deep transfer learning with joint adaptation networks.
\newblock In \emph{Proceedings of the 34th International Conference on Machine
  Learning-Volume 70}, pp.\  2208--2217. JMLR. org, 2017.

\bibitem[Long et~al.(2018)Long, Cao, Wang, and Jordan]{long2018conditional}
Long, M., Cao, Z., Wang, J., and Jordan, M.~I.
\newblock Conditional adversarial domain adaptation.
\newblock In \emph{Advances in Neural Information Processing Systems}, pp.\
  1640--1650, 2018.

\bibitem[Luo et~al.(2017)Luo, Zou, Hoffman, and Fei-Fei]{luo2017label}
Luo, Z., Zou, Y., Hoffman, J., and Fei-Fei, L.~F.
\newblock Label efficient learning of transferable representations acrosss
  domains and tasks.
\newblock In \emph{Advances in Neural Information Processing Systems}, pp.\
  165--177, 2017.

\bibitem[Nigam \& Ghani(2000)Nigam and Ghani]{nigam2000analyzing}
Nigam, K. and Ghani, R.
\newblock Analyzing the effectiveness and applicability of co-training.
\newblock In \emph{Cikm}, volume~5, pp.\ ~3, 2000.

\bibitem[Pan \& Yang(2009)Pan and Yang]{pan2009survey}
Pan, S.~J. and Yang, Q.
\newblock A survey on transfer learning.
\newblock \emph{IEEE Transactions on knowledge and data engineering},
  22\penalty0 (10):\penalty0 1345--1359, 2009.

\bibitem[Pan et~al.(2019)Pan, Yao, Li, Wang, Ngo, and
  Mei]{pan2019transferrable}
Pan, Y., Yao, T., Li, Y., Wang, Y., Ngo, C.-W., and Mei, T.
\newblock Transferrable prototypical networks for unsupervised domain
  adaptation.
\newblock In \emph{Proceedings of the IEEE Conference on Computer Vision and
  Pattern Recognition}, pp.\  2239--2247, 2019.

\bibitem[Pei et~al.(2018)Pei, Cao, Long, and Wang]{pei2018multi}
Pei, Z., Cao, Z., Long, M., and Wang, J.
\newblock Multi-adversarial domain adaptation.
\newblock In \emph{Thirty-Second AAAI Conference on Artificial Intelligence},
  2018.

\bibitem[Peng et~al.(2017)Peng, Usman, Kaushik, Hoffman, Wang, and
  Saenko]{peng2017visda}
Peng, X., Usman, B., Kaushik, N., Hoffman, J., Wang, D., and Saenko, K.
\newblock Visda: The visual domain adaptation challenge.
\newblock \emph{arXiv preprint arXiv:1710.06924}, 2017.

\bibitem[Pinheiro(2018)]{pinheiro2018unsupervised}
Pinheiro, P.~O.
\newblock Unsupervised domain adaptation with similarity learning.
\newblock In \emph{Proceedings of the IEEE Conference on Computer Vision and
  Pattern Recognition}, pp.\  8004--8013, 2018.

\bibitem[Quionero-Candela et~al.(2009)Quionero-Candela, Sugiyama, Schwaighofer,
  and Lawrence]{quionero2009dataset}
Quionero-Candela, J., Sugiyama, M., Schwaighofer, A., and Lawrence, N.~D.
\newblock \emph{Dataset shift in machine learning}.
\newblock The MIT Press, 2009.

\bibitem[Russakovsky et~al.(2015)Russakovsky, Deng, Su, Krause, Satheesh, Ma,
  Huang, Karpathy, Khosla, Bernstein, et~al.]{russakovsky2015imagenet}
Russakovsky, O., Deng, J., Su, H., Krause, J., Satheesh, S., Ma, S., Huang, Z.,
  Karpathy, A., Khosla, A., Bernstein, M., et~al.
\newblock Imagenet large scale visual recognition challenge.
\newblock \emph{International journal of computer vision}, 115\penalty0
  (3):\penalty0 211--252, 2015.

\bibitem[Saenko et~al.(2010)Saenko, Kulis, Fritz, and
  Darrell]{saenko2010adapting}
Saenko, K., Kulis, B., Fritz, M., and Darrell, T.
\newblock Adapting visual category models to new domains.
\newblock In \emph{European conference on computer vision}, pp.\  213--226.
  Springer, 2010.

\bibitem[Saito et~al.(2017)Saito, Ushiku, and Harada]{saito2017asymmetric}
Saito, K., Ushiku, Y., and Harada, T.
\newblock Asymmetric tri-training for unsupervised domain adaptation.
\newblock In \emph{Proceedings of the 34th International Conference on Machine
  Learning-Volume 70}, pp.\  2988--2997. JMLR. org, 2017.

\bibitem[Saito et~al.(2018)Saito, Watanabe, Ushiku, and
  Harada]{saito2018maximum}
Saito, K., Watanabe, K., Ushiku, Y., and Harada, T.
\newblock Maximum classifier discrepancy for unsupervised domain adaptation.
\newblock In \emph{Proceedings of the IEEE Conference on Computer Vision and
  Pattern Recognition}, pp.\  3723--3732, 2018.

\bibitem[Sankaranarayanan et~al.(2018)Sankaranarayanan, Balaji, Castillo, and
  Chellappa]{sankaranarayanan2018generate}
Sankaranarayanan, S., Balaji, Y., Castillo, C.~D., and Chellappa, R.
\newblock Generate to adapt: Aligning domains using generative adversarial
  networks.
\newblock In \emph{Proceedings of the IEEE Conference on Computer Vision and
  Pattern Recognition}, pp.\  8503--8512, 2018.

\bibitem[Shimodaira(2000)]{shimodaira2000improving}
Shimodaira, H.
\newblock Improving predictive inference under covariate shift by weighting the
  log-likelihood function.
\newblock \emph{Journal of statistical planning and inference}, 90\penalty0
  (2):\penalty0 227--244, 2000.

\bibitem[Shu et~al.(2018)Shu, Bui, Narui, and Ermon]{shu2018dirt}
Shu, R., Bui, H.~H., Narui, H., and Ermon, S.
\newblock A dirt-t approach to unsupervised domain adaptation.
\newblock \emph{arXiv preprint arXiv:1802.08735}, 2018.

\bibitem[Snell et~al.(2017)Snell, Swersky, and Zemel]{snell2017prototypical}
Snell, J., Swersky, K., and Zemel, R.
\newblock Prototypical networks for few-shot learning.
\newblock In \emph{Advances in Neural Information Processing Systems}, pp.\
  4077--4087, 2017.

\bibitem[Tan et~al.(2019)Tan, Peng, and Saenko]{tan2019generalized}
Tan, S., Peng, X., and Saenko, K.
\newblock Generalized domain adaptation with covariate and label shift
  co-alignment.
\newblock \emph{arXiv preprint arXiv:1910.10320}, 2019.

\bibitem[Torralba et~al.(2011)Torralba, Efros, et~al.]{torralba2011unbiased}
Torralba, A., Efros, A.~A., et~al.
\newblock Unbiased look at dataset bias.
\newblock In \emph{CVPR}, volume~1, pp.\ ~7. Citeseer, 2011.

\bibitem[Tsai et~al.(2018)Tsai, Hung, Schulter, Sohn, Yang, and
  Chandraker]{tsai2018learning}
Tsai, Y.-H., Hung, W.-C., Schulter, S., Sohn, K., Yang, M.-H., and Chandraker,
  M.
\newblock Learning to adapt structured output space for semantic segmentation.
\newblock In \emph{Proceedings of the IEEE Conference on Computer Vision and
  Pattern Recognition}, pp.\  7472--7481, 2018.

\bibitem[Tzeng et~al.(2014)Tzeng, Hoffman, Zhang, Saenko, and
  Darrell]{tzeng2014deep}
Tzeng, E., Hoffman, J., Zhang, N., Saenko, K., and Darrell, T.
\newblock Deep domain confusion: Maximizing for domain invariance.
\newblock \emph{arXiv preprint arXiv:1412.3474}, 2014.

\bibitem[Tzeng et~al.(2017)Tzeng, Hoffman, Saenko, and
  Darrell]{tzeng2017adversarial}
Tzeng, E., Hoffman, J., Saenko, K., and Darrell, T.
\newblock Adversarial discriminative domain adaptation.
\newblock In \emph{Proceedings of the IEEE Conference on Computer Vision and
  Pattern Recognition}, pp.\  7167--7176, 2017.

\bibitem[Venkateswara et~al.(2017)Venkateswara, Eusebio, Chakraborty, and
  Panchanathan]{venkateswara2017deep}
Venkateswara, H., Eusebio, J., Chakraborty, S., and Panchanathan, S.
\newblock Deep hashing network for unsupervised domain adaptation.
\newblock In \emph{Proceedings of the IEEE Conference on Computer Vision and
  Pattern Recognition}, pp.\  5018--5027, 2017.

\bibitem[Webb \& Ting(2005)Webb and Ting]{webb2005application}
Webb, G.~I. and Ting, K.~M.
\newblock On the application of roc analysis to predict classification
  performance under varying class distributions.
\newblock \emph{Machine learning}, 58\penalty0 (1):\penalty0 25--32, 2005.

\bibitem[Wen et~al.(2019)Wen, Zheng, Yuan, Gong, and Chen]{wen2019bayesian}
Wen, J., Zheng, N., Yuan, J., Gong, Z., and Chen, C.
\newblock Bayesian uncertainty matching for unsupervised domain adaptation.
\newblock \emph{arXiv preprint arXiv:1906.09693}, 2019.

\bibitem[Wu et~al.(2019)Wu, Winston, Kaushik, and Lipton]{wu2019domain}
Wu, Y., Winston, E., Kaushik, D., and Lipton, Z.
\newblock Domain adaptation with asymmetrically-relaxed distribution alignment.
\newblock \emph{arXiv preprint arXiv:1903.01689}, 2019.

\bibitem[Xie et~al.(2018)Xie, Zheng, Chen, and Chen]{xie2018learning}
Xie, S., Zheng, Z., Chen, L., and Chen, C.
\newblock Learning semantic representations for unsupervised domain adaptation.
\newblock In \emph{International Conference on Machine Learning}, pp.\
  5419--5428, 2018.

\bibitem[Zhang et~al.(2019{\natexlab{a}})Zhang, Zhang, Liu, and
  Tao]{zhang2019category}
Zhang, Q., Zhang, J., Liu, W., and Tao, D.
\newblock Category anchor-guided unsupervised domain adaptation for semantic
  segmentation.
\newblock In \emph{Advances in Neural Information Processing Systems}, pp.\
  433--443, 2019{\natexlab{a}}.

\bibitem[Zhang et~al.(2018)Zhang, Ouyang, Li, and Xu]{zhang2018collaborative}
Zhang, W., Ouyang, W., Li, W., and Xu, D.
\newblock Collaborative and adversarial network for unsupervised domain
  adaptation.
\newblock In \emph{Proceedings of the IEEE Conference on Computer Vision and
  Pattern Recognition}, pp.\  3801--3809, 2018.

\bibitem[Zhang et~al.(2019{\natexlab{b}})Zhang, Liu, Long, and
  Jordan]{pmlr-v97-zhang19i}
Zhang, Y., Liu, T., Long, M., and Jordan, M.
\newblock Bridging theory and algorithm for domain adaptation.
\newblock In Chaudhuri, K. and Salakhutdinov, R. (eds.), \emph{Proceedings of
  the 36th International Conference on Machine Learning}, volume~97 of
  \emph{Proceedings of Machine Learning Research}, pp.\  7404--7413, Long
  Beach, California, USA, 09--15 Jun 2019{\natexlab{b}}. PMLR.

\bibitem[Zou et~al.(2018)Zou, Yu, Vijaya~Kumar, and Wang]{zou2018unsupervised}
Zou, Y., Yu, Z., Vijaya~Kumar, B., and Wang, J.
\newblock Unsupervised domain adaptation for semantic segmentation via
  class-balanced self-training.
\newblock In \emph{Proceedings of the European Conference on Computer Vision
  (ECCV)}, pp.\  289--305, 2018.

\bibitem[Zou et~al.(2019)Zou, Yu, Liu, Kumar, and Wang]{Zou_2019_ICCV}
Zou, Y., Yu, Z., Liu, X., Kumar, B.~V., and Wang, J.
\newblock Confidence regularized self-training.
\newblock In \emph{The IEEE International Conference on Computer Vision
  (ICCV)}, October 2019.

\end{thebibliography}
\interlinepenalty=10000
\bibliographystyle{icml2020}

% for arxiv version
\newpage
\appendix
\section{Theory}
\begin{definition}
Let $\mathcal{B}_S$, $\mathcal{B}_T$ be minibatches from $\mathcal{U}_S$ and $\mathcal{U}_T$, respectively, where $\mathcal{B}_S\subseteq \mathcal{U}_S$, $\mathcal{B}_T\subseteq \mathcal{U}_T$, and $m_b=\vert \mathcal{B}_S\vert=\vert\mathcal{B}_T\vert$.
The empirical estimation of $d_{\mathcal{H}\Delta\mathcal{H}}(\mathcal{B}_S, \mathcal{B}_T)$ over the minibatches $\mathcal{B}_S$, $\mathcal{B}_T$ is defined as
\begin{align}
\hat{d}_{\mathcal{H}\Delta\mathcal{H}}(\mathcal{B}_S, \mathcal{B}_T)
=\frac{1}{m_b}\sup_{h,h'\in \mathcal{H}}\left \vert \sum_{\mathcal{B}_T}\left [ h\neq h' \right ]-  \sum_{\mathcal{B}_S}\left [ h\neq h' \right ]\right \vert.
\end{align}
\end{definition}
\vspace{-0.1in}
For simplicity, we drop the multiple $\frac{1}{m_b}$ in the following analysis as it does not affect the result of optimization.
\begin{theorem}[The decomposition of $\hat{d}_{\mathcal{H}\Delta\mathcal{H}}(\mathcal{B}_S, \mathcal{B}_T)$]
\label{theorem:bound}
Let $\mathcal{H}$ be a hypothesis space and $\mathcal{Y}$ be the label space of the classification task where $\mathcal{B}_S$, $\mathcal{B}_T$ are minibatches drawn from $\mathcal{U}_S$, $\mathcal{U}_T$, respectively, and $Y_S$, $Y_T$ are the label set of $\mathcal{B}_S$, $\mathcal{B}_T$. We define three disjoint sets on the label space: the shared labels $Y_C:=Y_S \cap  Y_T$, and the domain-specific labels $\overline{Y_{S}}:=Y_S-Y_C$, and $\overline{Y_{T}}:=Y_T-Y_C$.
We also define the following disjoint sets on the input space where $\mathcal{B}_S^C:=\left \{ x\in \mathcal{B}_S \mid y\in Y_C \right \}$, $\mathcal{B}^{\overline{C}}_{S}:=\left \{ x\in \mathcal{B}_S \mid y\notin Y_C \right \}$, $\mathcal{B}_T^C:=\left \{ x\in \mathcal{B}_T \mid y\in Y_C \right \}$, $\mathcal{B}^{\overline{C}}_{T}:=\left \{ x\in \mathcal{B}_T \mid y\notin Y_C \right \}$. 
The empirical $\hat{d}_{\mathcal{H}\Delta\mathcal{H}}(\mathcal{B}_S, \mathcal{B}_T)$ divergence can be decomposed into class aligned divergence and class-misaligned divergence:
\begin{equation}
    \hat{d}_{\mathcal{H}\Delta\mathcal{H}}(\mathcal{B}_S, \mathcal{B}_T)=\sup_{h,h'\in \mathcal{H}}\left |\xi^C(h, h') +  \xi^{\overline{C}}(h, h') \right |,
\end{equation}
where 
\begin{align}
    \xi^C(h, h')= \sum_{\mathcal{B}_T^C}\mathbbm 1\left [ h\neq h' \right ]-  \sum_{ \mathcal{B}_S^C}\mathbbm 1\left [ h\neq h' \right ],\\
    \xi^{\overline{C}}(h, h')= \sum_{\mathcal{B}^{\overline{C}}_{T}}\mathbbm 1\left [ h\neq h' \right ]-  \sum_{\mathcal{B}^{\overline{C}}_{S}}\mathbbm 1\left [ h\neq h' \right ].
\end{align}
\end{theorem}
\begin{proof}
By definition, we have
\begin{equation}
    \hat{d}_{\mathcal{H}\Delta\mathcal{H}}(\mathcal{B}_S, \mathcal{B}_T)=\sup_{h,h'\in \mathcal{H}}\left | \sum_{ \mathcal{B}_T}\mathbbm 1\left [ h\neq h' \right ]-  \sum_{\mathcal{B}_S}\mathbbm 1\left [ h\neq h' \right ] \right |
\end{equation}
We rewrite the summation over all the samples $\mathcal{B}$ into the sum of disjoint subsets $\mathcal{B}^C$ and $\mathcal{B}^{\overline{C}}$. 
\begin{align}
     &\sum_{\mathcal{B}_T}\mathbbm 1\left [ h\neq h' \right ]-  \sum_{\mathcal{B}_S}\mathbbm 1\left [ h\neq h'\right ]\\
     =& \left ( \sum_{\mathcal{B}_T^C}\mathbbm 1\left [ h\neq h' \right ] -  \sum_{\mathcal{B}_S^C}\mathbbm 1\left [ h\neq h'\right ] \right ) \\
     &+ \left ( \sum_{\mathcal{B}_T^{\overline{C}}}\mathbbm 1\left [ h\neq h' \right ] 
     - \sum_{\mathcal{B}_S^{\overline{C}}}\mathbbm 1\left [ h\neq h' \right ]\right ) \\
     =&\xi^C(h, h') +  \xi^{\overline{C}}(h, h').
\end{align}
This completes the proof.
\end{proof}

\section{Experiments}
\subsection{Additional Evaluation Measures on Office-Home}
\begin{table}[ht]
\renewcommand{\arraystretch}{1.15}
\centering
\caption{Evaluation on Office-Home~(\%) with ResNet-50.}
\label{tab:office-home-A2C-eval-measures}
\begin{center}
\scalebox{0.8}{
\begin{tabular}{lccccc}
\toprule
                   & \multicolumn{2}{c}{Ar$\shortrightarrow $Cl} &  & \multicolumn{2}{c}{Pr$\shortrightarrow $Rw} \\ \cline{2-3} \cline{5-6} 
                   & MDD             & ours           &  & MDD             & ours           \\ \midrule
accuracy           & 54.91               & 56.17               &  & 77.46               & 79.94               \\
macro F1 score     & 53.66               & 55.29               &  & 75.86               & 78.42               \\
weighted F1 score  & 53.97               & 55.81               &  & 77.24               & 79.79               \\
macro precision    & 57.02               & 57.72               &  & 78.21               & 79.56               \\
weighted precision & 58.85               & 60.30               &  & 79.60               & 80.97               \\
macro recall       & 56.41               & 57.76               &  & 76.30               & 78.61               \\
weighted recall    & 54.91               & 56.17               &  & 77.65               & 79.94               \\ \bottomrule
\end{tabular}
}
\end{center}
\end{table}
Table~\ref{tab:office-home-A2C-eval-measures} presents additional evaluation on Office-Home (standard).
We re-implement MDD using identical batch sizes~(50) and random seeds for fair comparison.
The results show that our proposed method has consistent improvements across all evaluation measures, and the improvements are not a result of batch sizes or random seeds.

\subsection{Additional Ablation on Alignment Options}
\begin{table}[ht]
\caption{The impact of different implicit alignment options, i.e., masking the classifier-based domain discrepancy measure and sampling examples from the source and target domains, on Ar$\rightarrow$Cl and Cl$\rightarrow$Pr, Office-Home~(standard).}
\renewcommand{\arraystretch}{1.2}
\centering
\scalebox{0.95}{
\begin{tabular}[!t]{cccc}
\toprule
 & \multicolumn{2}{c}{Alignment options} & \\ \cmidrule{2-3}
Domains                                 & masking               & sampling              & Accuracy \\\midrule
\multirow{4}{*}{Ar$\shortrightarrow$Cl} & $\times$   & $\times$    & 55.3                 \\
                                        & $\surd$   & $\times$    & 55.5                 \\
                                        & $\times$   & $\surd$    & 54.6                 \\
                                        & $\surd$   & $\surd$    & \textbf{56.2}                 \\\midrule
\multirow{4}{*}{Cl$\shortrightarrow$Pr} & $\times$   & $\times$    & 71.4                 \\
                                        & $\surd$   & $\times$    & 70.1                 \\
                                        & $\times$   & $\surd$    & 70.5                 \\
                                        & $\surd$   & $\surd$    & \textbf{73.1}       \\\bottomrule         

\end{tabular}
}
\label{tab:alignment-ablation-standard}
\end{table}
Table~\ref{tab:alignment-ablation-standard} presents the ablation study on Office-Home~(standard) that aims to assess the impact of different implicit alignment options: alignment in the domain divergence estimations~(i.e., \emph{masking} in MDD) and alignment in the input space~(i.e., \emph{sampling} class-conditioned examples).
We observe that both alignment techniques are essential for domain adaptation because alignment should be enforced consistently across all aspects of the domain adaptation training.
This is consistent to findings in the main paper.

\subsection{Learning Curve}
\label{sec:learning_curve}
\begin{figure}[ht]
    \centering
    \includegraphics[width=0.32\textwidth]{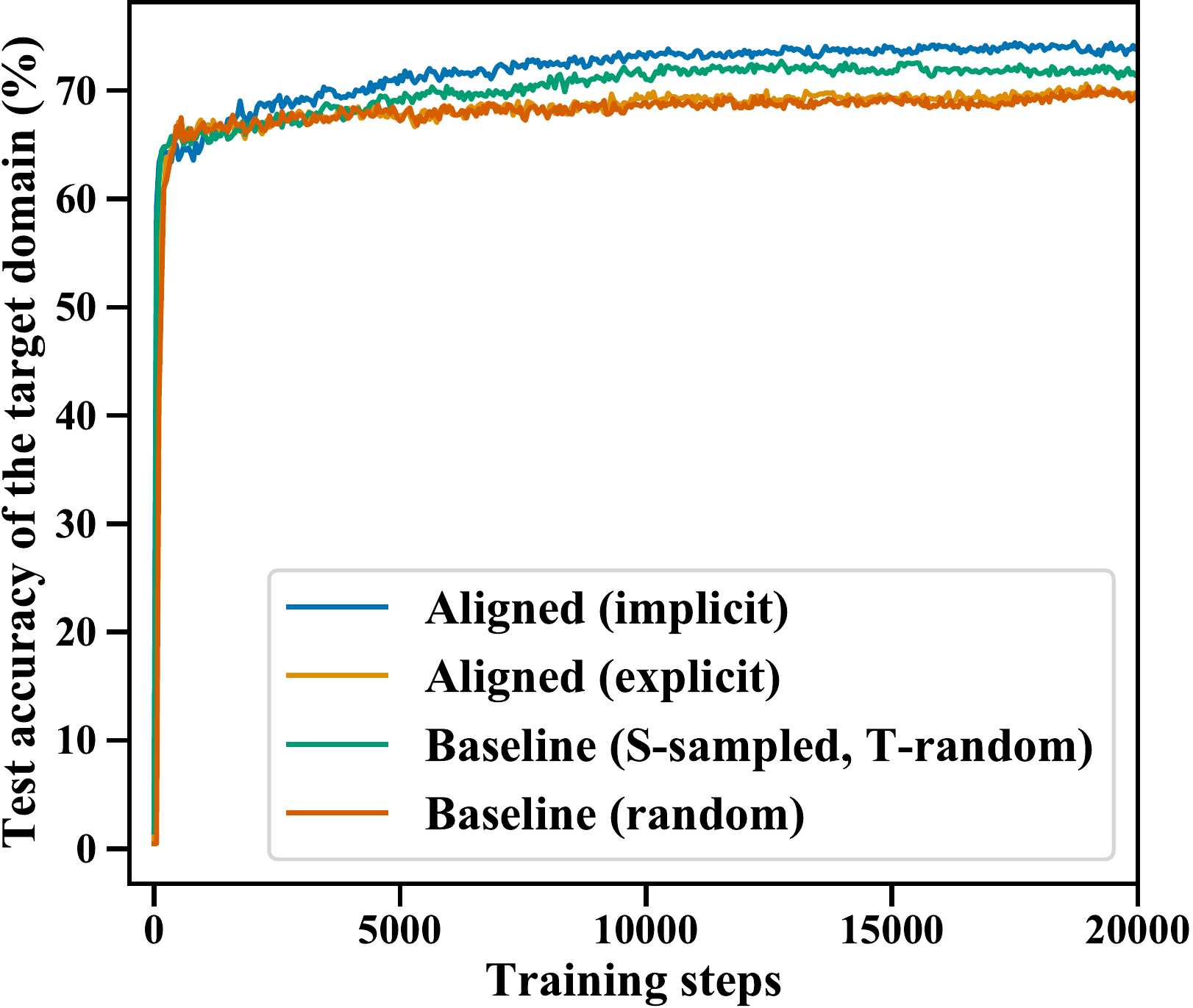}
    \caption{Learning curve of the target domain accuracy for Pr$\rightarrow$Rw, Office-Home~(RS-UT).}
    \label{fig:learning_curve_P2R}
\end{figure}
Figure~\ref{fig:learning_curve_P2R} shows the learning curve of the target domain accuracy for different methods.
The proposed implicit alignment converges better than other methods.
%We also find that explicit alignment has similar performances with the baseline, while implicit alignment performs the best.

\subsection{Computational Efficiency}
\begin{table}[ht]
\caption{The impact of pseudo-label update frequency on Ar$\rightarrow$Cl, Office-Home~(standard).}
\renewcommand{\arraystretch}{1.2}
\centering
\scalebox{0.9}{
\begin{tabular}{cc}
\toprule
pseudo-labels\\updated every $N$ steps & accuracy \\\midrule
5 & 56.0 \\
10 & 56.7 \\
20 & 56.2 \\
50 & 55.2 \\
100 & 56.3 \\
500 & 55.7 \\\bottomrule
\end{tabular}
}
\label{tab:update_freq}
\end{table}

Self-training requires estimating the target domain labels, which could be time-consuming depending on the size of the dataset.
To improve the computational efficiency of our algorithm, we only update pseudo-labels periodically, i.e., every 20 steps, instead of at every training step.
We show in Table~\ref{tab:update_freq} that different pseudo-label update frequencies exhibit similar performance on the target domain.
Notably, implicit alignment outperforms the baseline method in spite of only updating the pseudo-labels every 500 training steps.
This validates the robustness of implicit alignment.

For the experiments described in Section~\ref{sec:learning_curve}, training the baseline methods take 31 hours (wall clock time), whereas implicit alignment takes 34 hours under the same training condition when the pseudo-labels are updated every 20 steps.
The 10\% computational overhead is rather restricted.
Moreover, from an engineering perspective, partially updating and caching the pseudo-labels could further improve the computational efficiency, and we leave them as future work.

\subsection{Impact of Batch Size}
\begin{table}[h]
\caption{Impact of batch size on target domain accuracy~(\%), Ar$\rightarrow$Cl, Office-Home~(standard). The MDD results are based on our re-implementation.}
\renewcommand{\arraystretch}{1.2}
\centering
\scalebox{0.95}{
\begin{tabular}{ccc}
\toprule
batch size & baseline & implicit \\\midrule
8 & 48.9 & 49.7 \\
16 & 52.7 & 52.8 \\
32 & 54.9 & 56.2 \\
50 & 55.3 & 56.2 \\\bottomrule
\end{tabular}
}
\label{tab:batchsize_ablation}
\end{table}
Table~\ref{tab:batchsize_ablation} presents the impact of batch size on the target domain accuracy.
We find that implicit alignment consistently improves the model performance over the MDD baseline across different batch sizes, and both methods work better with larger batch sizes.

\subsection{Empirical Class Diversity}
\begin{figure}[h]
    \centering
    \includegraphics[width=0.33\textwidth]{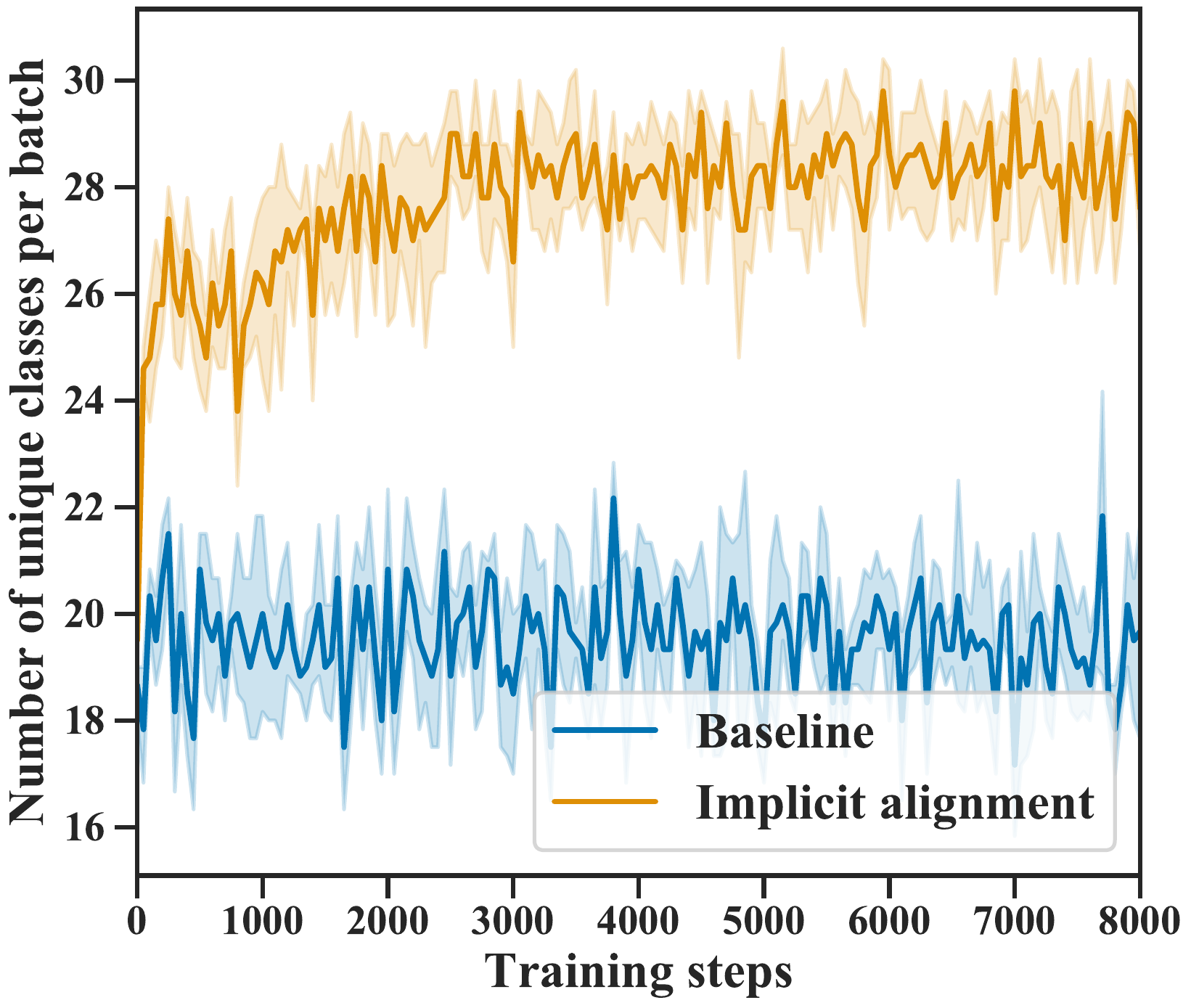}
    \caption{Empirical class diversity while training A$\rightarrow$W (Office-31) with batch size 31.}
    \label{fig:empirical_distribution_of_alignment}
\end{figure}
Figure~\ref{fig:empirical_distribution_of_alignment} shows the empirical class diversity comparing implicit alignment with the baseline.
In both experiments, the batch size is identical with the total number of classes~(i.e., 31).
For the baseline method, random sampling only obtains about 19 unique classes per-batch, which is much smaller than the batch size, in spite of the batch sizes being the same with the total number of classes.
This is because random sampling can be viewed as sampling \emph{with replacement} in the label space, whereas the implicit alignment can be viewed as sampling \emph{without replacement} in the label space, which naturally increases the empirical class diversity.
The expected class diversity of the baseline is
\begin{equation}
    \E\left [ \left | Y \right | \right ]=n\left [ 1-\left ( \frac{n-1}{n} \right )^k \right ],
\end{equation}
where $n$ is the number of unique classes and $k$ is the size of the minibatch.
The expected class diversity is 19.78 if $n=31$ and $k=31$, which is consistent with the empirical class diversity shown in Figure~\ref{fig:empirical_distribution_of_alignment}.

For the implicit alignment method shown in Figure~\ref{fig:empirical_distribution_of_alignment}, although it has low class diversity at training step 0 due to the random pseudo-labels, it has a sharp increase in class diversity for the first few hundred training steps, and eventually being able to sample 28 classes from the total of 31 classes.
This confirms that implicit alignment is effective in improving empirical class diversity beyond random sampling.

\section{Datasets}
\begin{figure}[h]
    \centering
    \includegraphics[width=0.5\textwidth]{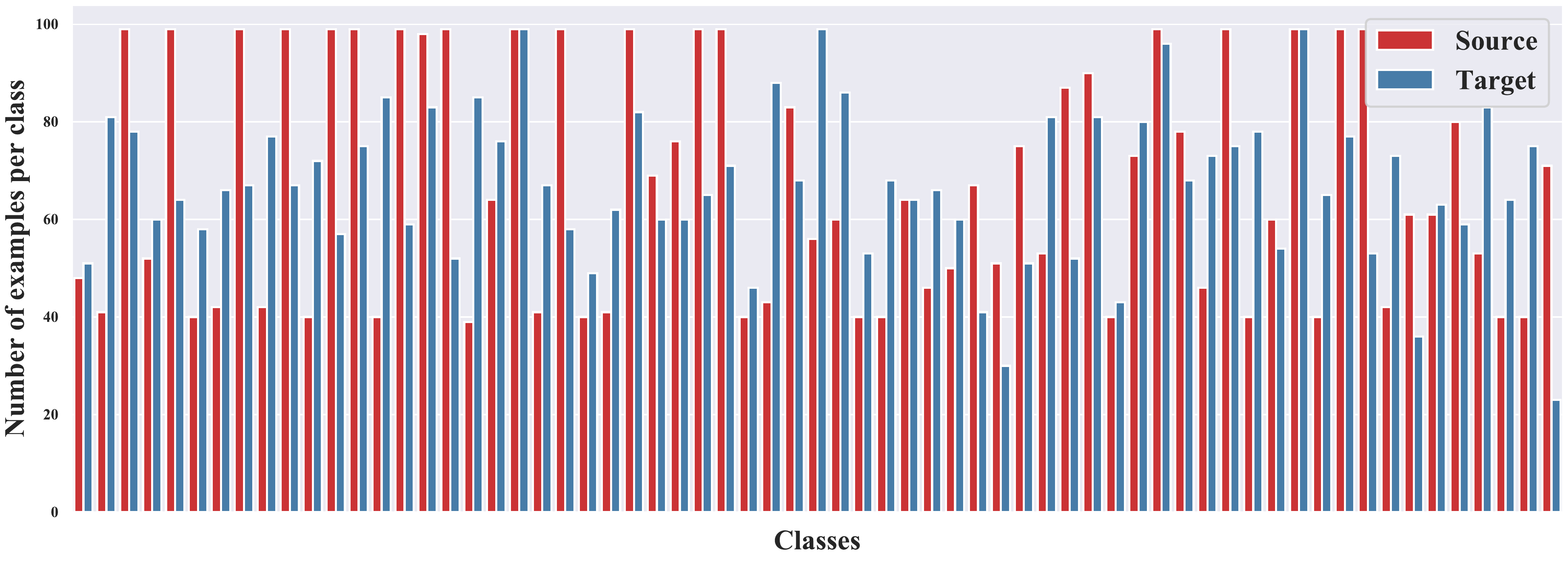}
    \caption{Class frequency of Cl$\rightarrow$Rw, Office-Home~(standard)}
    \label{fig:class-distribution-office-home-standard}
\end{figure}

Figure~\ref{fig:class-distribution-office-home-standard} shows the frequencies of different classes for Cl$\rightarrow$Rw on the Office-Home~(standard) dataset.
This dataset is under natrual class imbalance where examples of different classes are not evenly distributed.

\begin{figure}[h]
    \centering
    \includegraphics[width=0.5\textwidth]{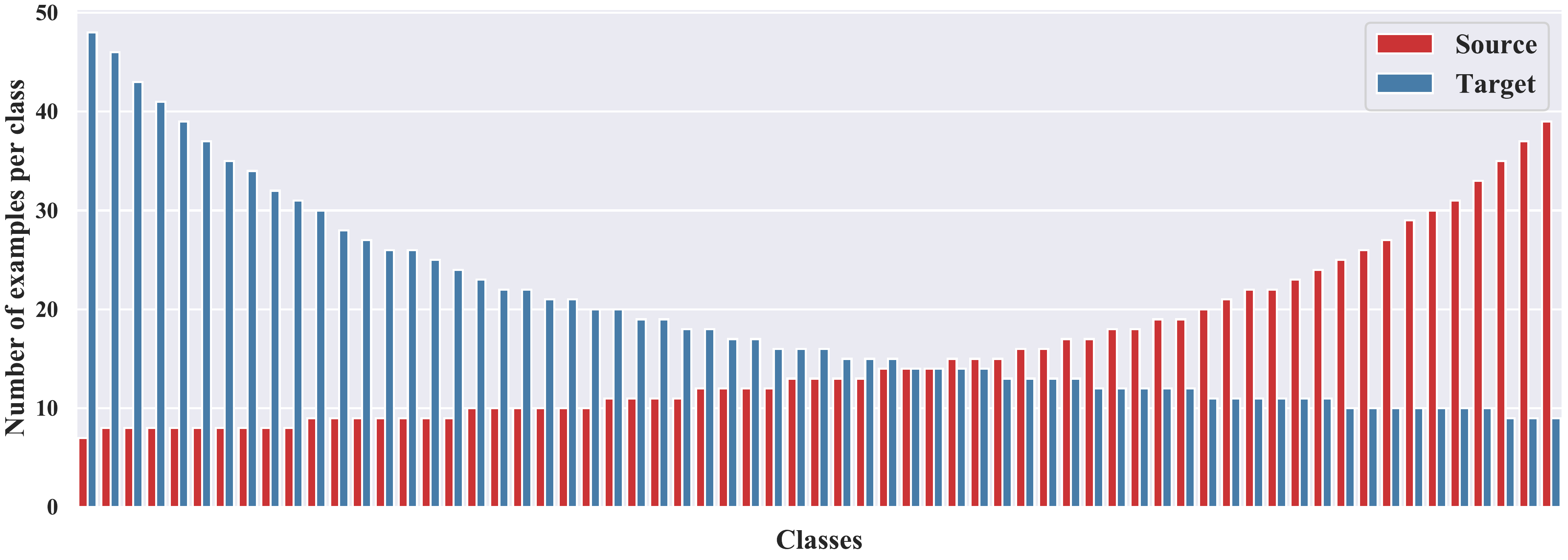}
    \caption{Class distribution of of Cl$\rightarrow$Rw, Office-Home~(RS-UT)}
    \label{fig:class-distribution-office-home-rs-ut}
\end{figure}

Figure~\ref{fig:class-distribution-office-home-rs-ut} shows the frequencies of different classes for Cl$\rightarrow$Rw on the Office-Home~(RS-UT) dataset~\cite{tan2019generalized}.
In this dataset, the minority classes in the source domain are majority classes in the target domain, which creates extreme within-domain class imbalance and between-domain distribution shift.

\section{Model Architecture and Training Details}
\paragraph{Code.}~We use PyTorch 1.2 as the training environment, and we observe that the adaptation performance on PyTorch 1.4 is slightly better PyTorch 1.2.
The differences between PyTorch versions do not change the findings and the conclusions of this paper.
Our code and training instructions are provided in \url{https://github.com/xiangdal/implicit_alignment}.

\paragraph{Model architecture.} We use ResNet-50~\cite{he2016deep} pre-trained from ImageNet~\cite{russakovsky2015imagenet} as the backbone, and use hyper-parameters from~\cite{pmlr-v97-zhang19i} for MDD-based domain discrepancy measure.
The backbone is followed by a 1-1ayer bottleneck.
The classifier $f$ and auxiliary classifier $f'$ are both 2-layer networks.

\paragraph{Optimization.} We use the SGD optimizer with learning rate $0.001$, nesterov momentum $0.9$, and weight decay $0.0005$.
We empirically find that SGD converges better than Adam for adversarial optimization.
We use a gradient reversal layer for minimax optimization, and we use a training scheduler~\cite{ganin2016domain} for gradient reversal layer defined as
\begin{equation}
    \lambda_p=\frac{0.2}{1+\mathrm{exp}(-\frac{i}{1000})} - 0.1,
\end{equation}
where $i$ denotes the step number.
We used the same scheduler from~\cite{pmlr-v97-zhang19i} for all experiments and have not tried hyperparameter search for $\lambda_p$.
The batch size is 31 for Office-31 and 50 for Office-Home.

\end{document}